\documentclass{article} 
\usepackage{iclr2025_conference,times}

\usepackage{amsmath,amsfonts,bm}

\def\eqref#1{equation~\ref{#1}}

\def\1{\bm{1}}

\DeclareMathAlphabet{\mathsfit}{\encodingdefault}{\sfdefault}{m}{sl}
\SetMathAlphabet{\mathsfit}{bold}{\encodingdefault}{\sfdefault}{bx}{n}

\usepackage{hyperref}
\usepackage{url}
\usepackage{natbib}
\usepackage[capitalise]{cleveref}
\usepackage{multirow}
\usepackage{graphicx}
\usepackage{subfigure}
\usepackage{booktabs}
\usepackage{siunitx}
\usepackage{wrapfig}
\usepackage{subcaption}
\usepackage{subfigure}
\usepackage{array}
\usepackage{longtable}
\usepackage{amssymb}
\usepackage{enumitem}
\usepackage{pifont}
\usepackage{xcolor}

\title{No Location Left Behind: Measuring and Improving the Fairness of Implicit Representations for Earth Data}

\author{Daniel Cai, Randall Balestriero \\
Department of Computer Science\\
Brown University\\
}

\iclrfinalcopy 
\begin{document}

\maketitle

\begin{abstract}
Implicit neural representations (INRs) exhibit growing promise in addressing Earth representation challenges, ranging from emissions monitoring to climate modeling. However, existing methods disproportionately prioritize global average performance, whereas practitioners require fine-grained insights to understand biases and variations in these models.
To bridge this gap, we introduce \textsc{FAIR-Earth}: a first-of-its-kind dataset explicitly crafted to examine and challenge inequities in Earth representations. \textsc{FAIR-Earth} comprises various high-resolution Earth signals and uniquely aggregates extensive metadata along stratifications like landmass size and population density to assess the fairness of models.
Evaluating state-of-the-art INRs across the various modalities of \textsc{FAIR-Earth}, we uncover striking performance disparities. Certain subgroups, especially those associated with high-frequency signals (e.g., islands, coastlines), are consistently poorly modeled by existing methods.
In response, we propose spherical wavelet encodings, building on previous spatial encoding research. Leveraging the multi-resolution capabilities of wavelets, our encodings yield consistent performance over various scales and locations, offering more accurate and robust representations of the biased subgroups. These open-source contributions\footnote{Available as part of the \textsc{FAIR-Earth} package (\href{https://github.com/raccooncai448/FAIREarth}{\texttt{github.com/raccooncai448/FAIREarth}})} represent a crucial step towards the equitable assessment and deployment of Earth INRs.
\end{abstract}

\begin{figure}[h!]
    \centering
    \vspace{-0.5cm}
    \includegraphics[width=0.95\linewidth]{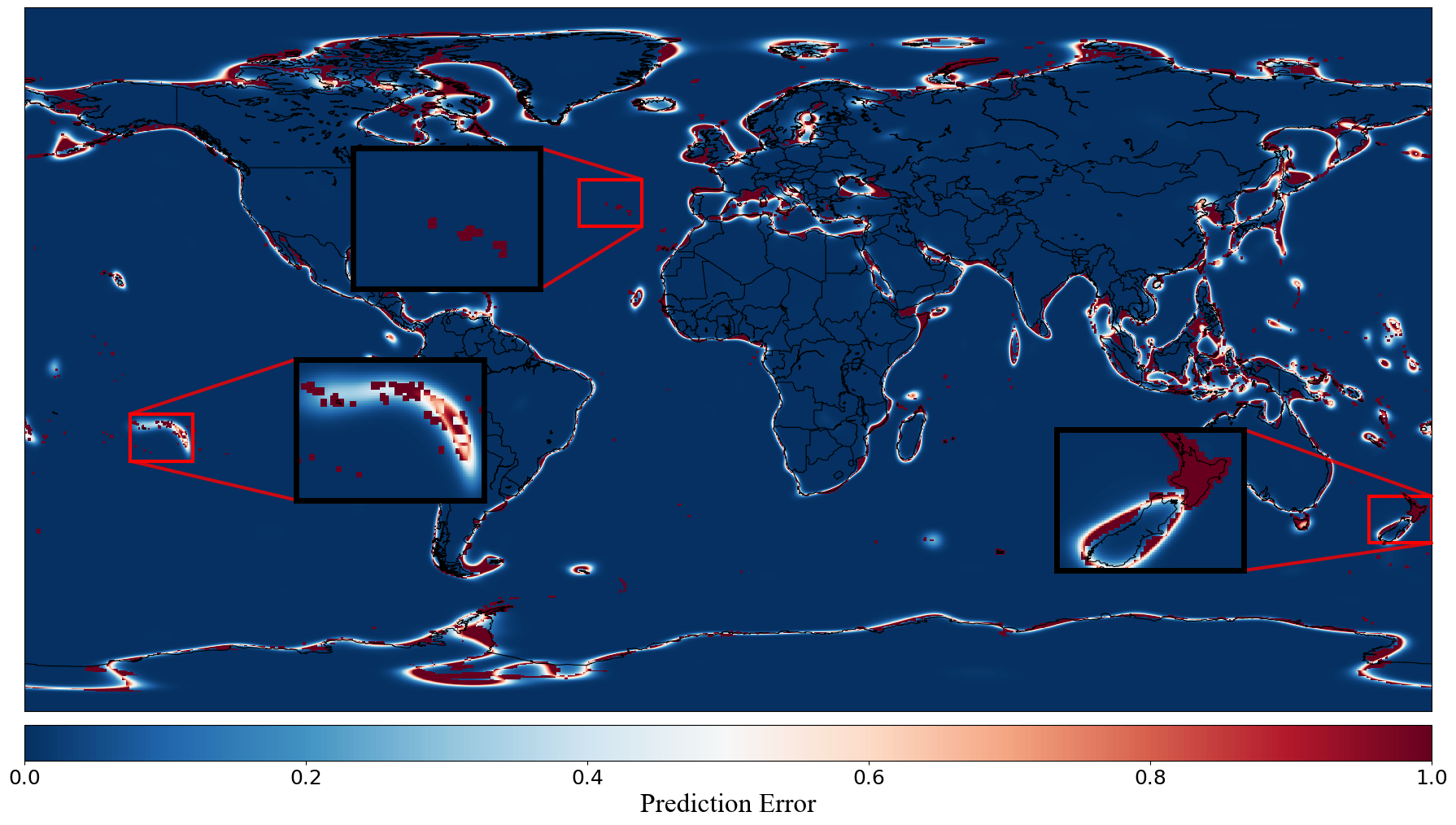}
    \vspace{-0.25cm}
    \caption{\small Heatmap of the spatial distribution of approximation errors using existing INRs to model land-sea data of the Earth. Leveraging the resolution of \textsc{FAIR-Earth}, {\bf we uncover clear bias against islands, where error magnitude is significantly higher}. 
    Against the same task, \textsc{Spherical Wavelets} resolves these issues by \textbf{reconciling global signals with fine and localized signals}. Details and plots available in \cref{APP:Misc Figures}.
    }
    \label{fig:failure_cases}
\end{figure}

\section{Introduction}

Implicit neural representations (INRs) have experienced a rapid surge in popularity since the groundbreaking work of \citet{mildenhall2020nerfrepresentingscenesneural} first demonstrated their effectiveness in mapping complex 3D scenes to continuous functions. Fundamentally, early INRs were formulated as a function $f_\theta: X \rightarrow Y$ with sinusoidal activation and weights $\theta$, where $X$ is typically a coordinate space and $Y$ is the target space \citep{sitzmann2020implicit}. The key enabling feature of this representation was the continuous support of $f_\theta$, yielding resolution-independent representations.

In recent years, the versatility and advantages of INRs have made them increasingly attractive across diverse fields \citep{molaei2023implicitneuralrepresentationmedical, chen2023implicitneuralspatialrepresentations, rußwurm2024geographiclocationencodingspherical}. Unlike traditional discrete methods, INRs are essentially infinite-resolution models, enabling them to effectively handle signal-processing tasks involving sharp discontinuities or high variance features \citep{xu2022signalprocessingimplicitneural}. Furthermore, they naturally excel in scenarios requiring differentiable representations, facilitating tasks such as gradient-based optimization in physics-informed machine learning \citep{raissi2017physicsinformeddeeplearning}. The compact and flexible parameterization of INRs also offers substantial memory savings compared to grid-based methods, a critical consideration for any practitioner working with resource constraints \citep{liu2024efficientimplicitneuralrepresentation}.

More recently, INRs have made notable strides in Earth science applications, with transformative potential for pressing tasks such as climate modeling and environmental monitoring \citep{rußwurm2024geographiclocationencodingspherical, bookSeismic}. While existing methods span various techniques such as observation-based networks \citep{palecki2013uscrn}, satellite-based remote sensing \citep{sorooshian2014persiann}, and more recently, computer-based climate simulations \citep{geneva_foster_2024_earth2studio}, these  methods suffer from some combination of discretization error, data inconsistency, and resource-intensive inference \citep{allen2002model}. In contrast, the ability of INRs to learn nonparametric models from arbitrarily high-resolution, multi-modal data presents an efficient and promising alternative. Specifically, the use of INRs to learn the underlying dynamics of explicitly \textit{geospatial data} has rapidly garnered interest \citep{cole2023spatial}. These methods encode the generative process and representation of the data through an implicit function, typically a Deep Neural Network (DNN), that maps spatial coordinates to data realizations \citep{hillier2023geoinr}.

However, as INRs transition into these realms of Earth science application, it is imperative to thoroughly evaluate the fairness and potential biases of these models. Skewed or unfair models in this domain can have far-reaching consequences, potentially amplifying existing societal inequalities or misallocating resources in vulnerable communities \citep{Munday2018, Flores2022}. For instance, two concrete instances of modeling bias in the Earth science and INR domains underscore the critical importance of fairness considerations:
\begin{itemize}
    \item Federal Emergency Management Agency (FEMA) flood maps, which rely on an ensemble of machine learning models for risk assessment, have been shown to systematically underestimate flood risk in lower-income neighborhoods \citep{Flores2022}. 
    \item Global climate models, including those used in the Coupled Model Intercomparison Project (CMIP), exhibit systematic errors in simulating precipitation patterns over Africa, consistently overestimating precipitation in southern regions while underestimating it in central regions \citep{Munday2018}.
\end{itemize}
While such fairness concerns should evidently be top-of-mind, as of writing, there exist \textit{no systematic datasets} to target such biases for implicit Earth representations. Hence, we introduce the first-of-its-kind \textsc{FAIR-Earth} (Fairness Assessment for Implicit Representations of Earth Data) dataset, a comprehensive framework for evaluating and mitigating biases of implicit Earth representations. 

\begin{figure}[t!]
    \centering
    \includegraphics[width=\textwidth]{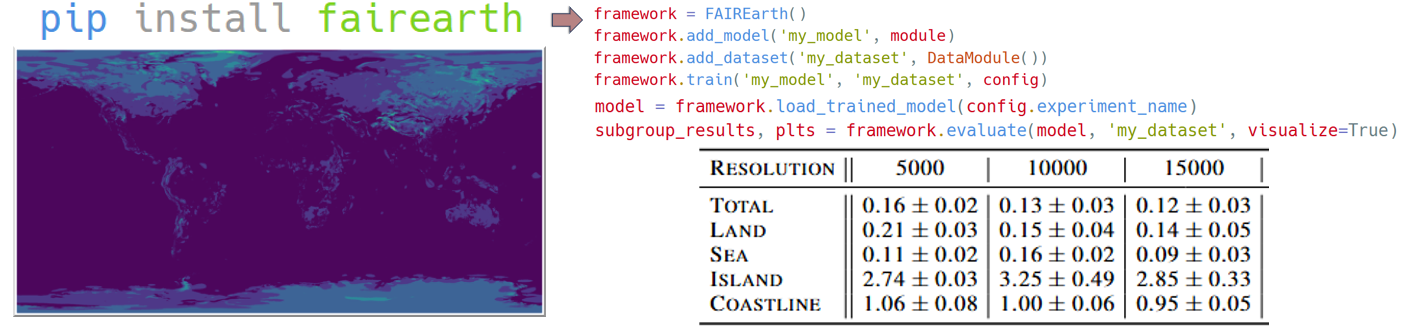}
    \caption{\textsc{FAIR-Earth} is the first Earth INR-framework for fairness assessment, and moreover is an \textit{open-source package} that allows for agile testing and analysis of subgroup-level performance.}
    \label{fig:codeFig}
\end{figure}

{\bf Contribution 1:}~\textsc{FAIR-Earth} comprises a diverse set of benchmark datasets representing a wide spectrum of Earth science signals, from global precipitation to carbon emissions (\cref{FAIREarth}). Uniquely, it also includes a suite of methodologies and metadata designed to quantify disparities in model performance across different stratifications, setting it apart from existing datasets (\cref{tab:dataset_comparison}). In essence, \textsc{FAIR-Earth} unifies disparate existing geospatial datasets under a consistent framework, enriched with extensive metadata spanning fields such as geographical features (e.g., islands, coastlines), demographic information (e.g., population density) and geopolitical boundaries (e.g., country divisions). 

Extensive experiments against \textsc{FAIR-Earth} reveal significant disparities in the performance of state-of-the-art INR methods across different subgroups (\cref{Evaluation}). In particular, our analysis uncovers a strong negative correlation between landmass size and representation loss, with areas corresponding to localized signals and high-frequency features exhibiting consistently poor performance (\cref{fig:sidebyside}). 

To address these fairness issues, we build upon recent research into \textit{location encodings}, which aim to simplify downstream Earth representation tasks by transforming the input into a more \textit{learning-friendly} format. Previous work in this domain \citep{rußwurm2024geographiclocationencodingspherical} has focused on improving location encodings via decomposition onto the harmonic domain. While these Fourier-inspired approaches have demonstrated improvements in global accuracy and longitudinal consistency, our empirical results show that the global support of Fourier bases introduces biases against localized signals, compromising more fine-scaled geographic representations (\cref{fig:failure_cases}).

{\bf Contribution 2:}~With these limitations in mind, we propose \textsc{Spherical Wavelets} (SW), a novel encoding mechanism grounded in existing research in the wavelet domain (\cref{SW}). By explicitly modeling geospatial phenomena at multiple scales, \textsc{Spherical Wavelets} enable efficient and accurate modeling of regions with highly localized signals (\cref{fig:SWprediction}). Our approach is motivated by the inherent multi-scale nature of Earth systems, and allows for more equitable representation across diverse geographical contexts. Against the \textsc{FAIR-Earth} framework, we demonstrate that our encodings significantly reduce performance disparities compared to existing methods, paving the way for fairer Earth science applications (\cref{fig:visual_comparison}).

We summarize our contributions below:
\begin{itemize}
    \item We introduce \textsc{FAIR-Earth}, a comprehensive framework for assessing fairness of implicit Earth representations across multiple modalities, enabling users to perform any fine-grained fairness study as required by their application (\cref{FAIREarth}).
    \item We perform a thorough fairness evaluation of current INR frameworks against \textsc{FAIR-Earth}. We discover significant subgroup disparities in current state-of-the-art INR methods (\cref{Evaluation}).
    \item We propose spherical wavelet encodings for INRs, motivated by the shortcomings of existing INRs when dealing with multi-resolution data. Against \textsc{FAIR-Earth}, we demonstrate competitive performance while significantly mitigating per-group biases (\cref{sw-biases}).
\end{itemize}
We hope that our study will pave the way towards more fair and equitable implicit Earth representations.

\section{Background}
\label{Background}
{\bf Fairness of Implicit Neural Representations}~
Recent work has shown that optimizing solely for average test performance can have detrimental effects on sub-group performance in natural image classification tasks \citep{balestriero2022effectsregularizationdataaugmentation, kirichenko2023understandingdetrimentalclassleveleffects}. These findings raise natural concerns about potential fairness issues in implicit neural representations (INRs), particularly in the context of Earth data. For example, algorithms fine-tuned to low-frequency global signals may overlook high-frequency fluctuations, potentially leading to biased representations. However, to the best of our knowledge, a \textit{systematic study of fairness} in INRs for Earth data has yet to be conducted.

We focus on the work by \citet{rußwurm2024geographiclocationencodingspherical}, which set state-of-the-art benchmark performance by decomposing positional information onto spherical harmonic basis functions and integrating these encodings with SirenNets \citep{sitzmann2020implicit}. The core contribution is the representation of Earth data as continuous signals on the globe $f: (\lambda, \phi) \mapsto \mathbb{R}$, where $\lambda$ and $\phi$ denote longitude and latitude, respectively. For well-behaved functions with exponential decay of their eigenvalues, the signal can be precisely recovered using the following decomposition:

\begin{align}
f(\lambda, \phi) = \sum_{l=0}^{\infty} \sum_{m=-l}^{l} w_l^m Y_l^m(\lambda, \phi),    \label{eq:decomposition}
\end{align}

where $Y$ is the class of spherical harmonic functions, $w$ are learnable scalar weights, and $l$ and $m$ are the degree and order of the basis function $Y_l^m$. 
Both as a practice and a necessity (\cref{fig:legendreNans}), an upper bound on $l$ is imposed, restricting the embedding size to $l^2$ and capping the representable frequency.


{\bf Location Encodings for INRs}~
\label{background-locationencoding}
The work in \citet{rußwurm2024geographiclocationencodingspherical} notably represents a significant advancement in the task of constructing proper \textit{location encodings} for implicit Earth representations. In a general sense, location encodings can be viewed as a simpler alternative to embeddings, a familiar concept in natural language or multimodal learning fields. Like embeddings, encodings play a crucial role in transforming the complex task of learning the distribution $P(y|\textbf{x})$ into a simpler task $P(y|\textbf{x*})$, where $\textbf{x}$ is the location, $y$ is the predictive task, and $\textbf{x*}$ is the location encoding \citep{Mai_2022}. 

\begin{figure}[t!]
    \centering
    \subfigure[\textsc{Cartesian3D} \citep{tseng2022timltaskinformedmetalearningagriculture}]{
        \includegraphics[width=0.22\textwidth]{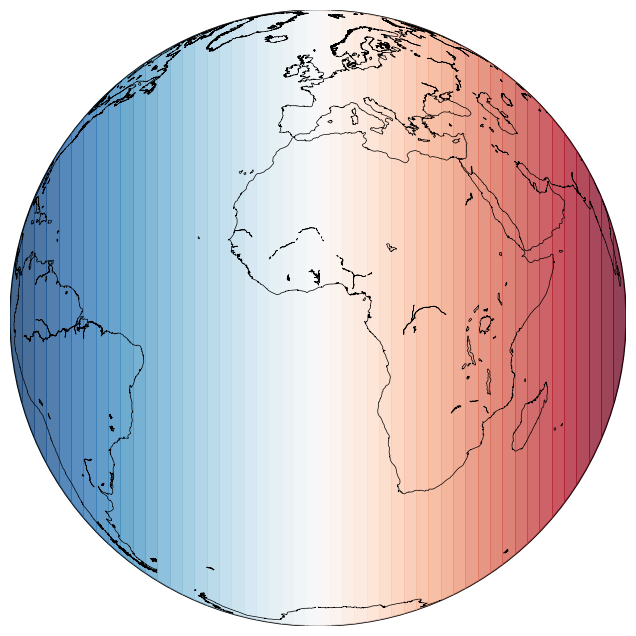}
    }
    \subfigure[\textsc{Theory} \citep{mai2020multiscalerepresentationlearningspatial}]{
        \includegraphics[width=0.22\textwidth]{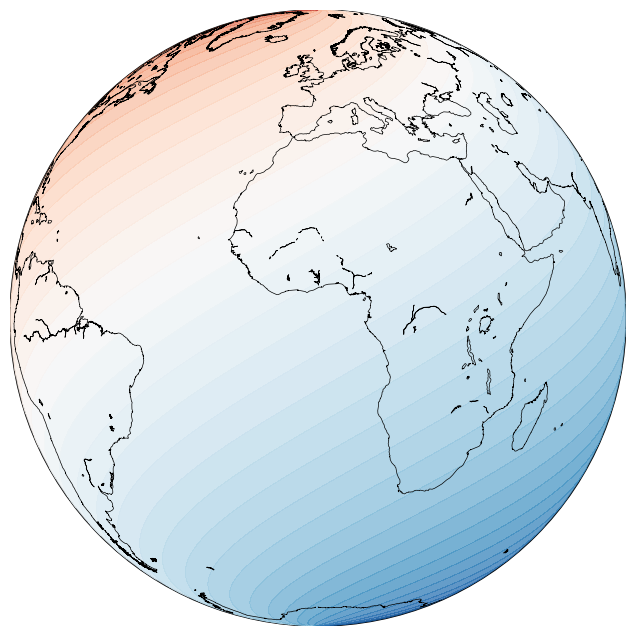}
    }
    \subfigure[\textsc{Grid and Sphere} \citep{mai2023sphere2vecgeneralpurposelocationrepresentation}]{
        \includegraphics[width=0.22\textwidth]{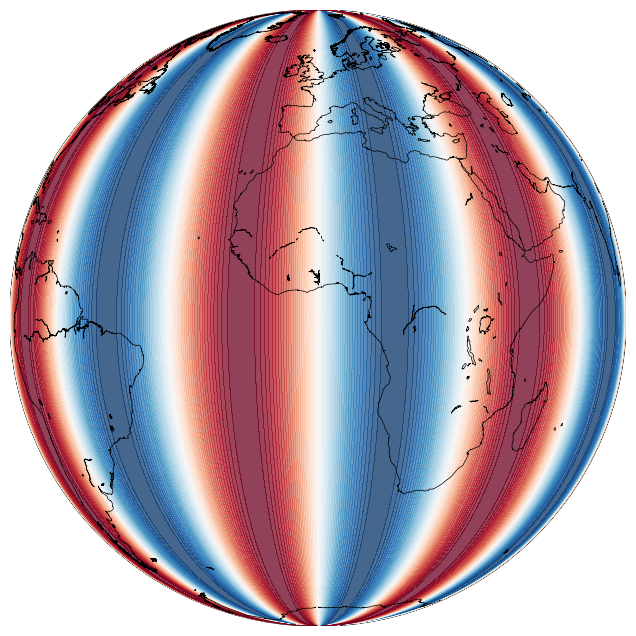}
    }
    \subfigure[\textsc{Spherical Harmonic} \citep{rußwurm2024geographiclocationencodingspherical}]{
        \includegraphics[width=0.22\textwidth]{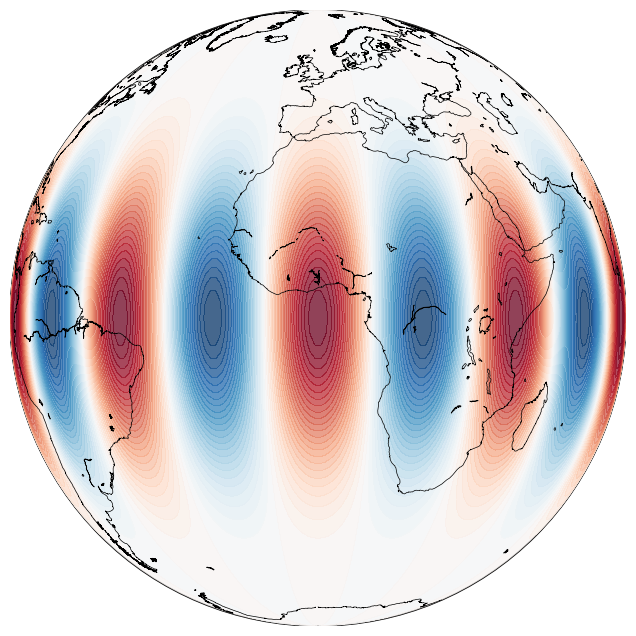}
    }
    \caption{Visualization of existing encoding mechanisms.}
    \label{fig:filterbanks}
    \vspace{-4mm}
\end{figure}

Early location encodings leveraged Gaussian processes and kernel methods \citep{Mai_2022}, while more recent approaches have improved performance even with simpler mappings like \textsc{Direct} ($(\phi, \lambda) \mapsto (\phi, \lambda)$) and \textsc{Cartesian3D} ($(\phi, \lambda) \mapsto (x, y, z)$). As datasets have grown in dimensionality, the field has built upon variants of \textsc{Grid}/\textsc{Theory} embeddings, which involve simple interaction terms between sine and and cosine functions on the globe. Notable examples include the \textsc{Grid and Sphere} family of encodings (\textsc{SphereC+} and \textsc{SphereM+}) which generalizes and combines these previous encodings (\cite{mai2023sphere2vecgeneralpurposelocationrepresentation}). More recently, as INRs have gained popularity, the focus has shifted towards more complex periodic representations of geospatial data, as exemplified by the harmonic-based encodings in \citet{rußwurm2024geographiclocationencodingspherical}. 

However, these global decomposition methods (\cref{tab:existing_encodings_comparison}) may not be accurate or efficient for common natural signals across the globe (\cref{fig:filterbanks}). Fourier and harmonic-based methods are especially known to alias in localized, sharp, or discontinuous portions of the signal \citep{mallat1999wavelet}, which intuitively can lead to biased representations of important Earth features. 

{\bf Measuring Fairness}~
As described, the global approach adopted by many existing implicit Earth representations is problematic when dealing with naturally localized or discontinuous signals. Before attempting to mitigate these biases, it is crucial for practitioners to understand their nature and extent. More specifically, there is a pressing need to quantify the trade-offs in per-group performance of existing models \citep{garciasilva2018enablingfairresearchearth}. This emphasis on group-specific performance is not merely an academic exercise but is driven by practical considerations: for tasks like natural disaster risk assessment where consequences are severe, there is a natural emphasis on improving worst-case performance rather than average-case metrics \citep{kemp2022climate}.

An ideal framework for measuring fairness in Earth representations should encompass several key components: consistent high-resolution data for uniform assessment, multi-modal integration to reflect various Earth systems, and standardized fairness metrics to quantify biases along various themes. By developing such a comprehensive framework, the implicit learning and Earth science communities can work towards constructing more equitable models, aligning themselves with the growing emphasis on fairness and accountability in AI systems.

\section{FAIR-Earth: All-in-One Evaluation Library for Earth Data}

The introduction of \textsc{FAIR-Earth}, the first-of-its-kind dataset to target per-group performance, is highly motivated by both the natural tendencies of geospatial data and the well-studied biases of existing INRs. We first examine the specific inadequacies of existing datasets, and then outline the main contributions of the \textsc{FAIR-Earth} dataset.

\subsection{The Need for a Fairness Dataset}
\label{sec:motivation}

The current landscape of Earth system datasets is not well-suited for the nuanced analyses described earlier. Commonly used datasets and benchmarks suffer from inherent limitations that can introduce or exacerbate biases. We point to two concrete examples below.

{\bf The Shuttle Radar Topography Mission}~(SRTM) land elevation model, despite its widespread use, suffers significant data gaps and inconsistencies in areas with steep terrain or dense vegetation \citep{SRTM2013}. These gaps can lead to skewed representations of topographical features, potentially affecting critical applications such as flood risk assessment or infrastructure planning in vulnerable areas.


{\bf The Visible Infrared Imaging Radiometer Suite}~(VIIRS) nighttime lights dataset, often employed as a proxy for economic activity and urbanization, has limitations in detecting low levels of light emissions \citep{Murphy2006}. This shortcoming may  underestimate activity in rural or underdeveloped areas, reinforcing existing biases in policy decisions.

These examples underscore a critical gap in the field: while existing datasets and Earth system models are clearly at risk of exhibiting biases across different stratifications, there is no established framework specifically designed to quantify these biases. 
\subsection{Proposed Evaluation Dataset}
\label{FAIREarth}
To mitigate the limitations of some data sources highlighted in \cref{sec:motivation}, \textsc{FAIR-Earth} is based around several clean, accessible, and interpretable data modalities. Additionally, we collect a comprehensive suite of metadata to assess biases that may arise from both the data and representation model; this metadata lays the crucial groundwork for our novel fairness assessment of implicit Earth representations.

{\bf Land-Sea Boundaries.}~Based on NASA's Integrated Multi-satellitE Retrievals for GPM (IMERG) dataset \citep{huffman2014imerg}, this component contains (\cref{fig:image1}) coarse signals like continental landmasses while also providing high-resolution boundaries for fine-grained signals such as islands and coastlines.

{\bf CO2 Emissions.} Using data primarily from NASA's Orbiting Carbon Observatory-2 (OCO-2) satellite, this component provides ultra fine-grained and precise information into carbon emissions \citep{OCO2, ODell2018, Taylor2023}. OCO-2 is the second high-precision CO2 satellite, and inaccuracies due to reflectance or cloud cover are mitigated via data assimilation. This high-resolution emissions data (\cref{fig:image8}) is crucial for environmental justice applications, for instance enabling researchers to identify localized pollution hotspots, or investigate correlations between emissions patterns and health disparities. 

{\bf Precipitation and Temperature.} Derived from the CHELSA \citep{karger2017climatologies} dataset, this component provides an assortment of coarse and ultra fine-grained signals, as well as high resolution along the time dimension (\cref{fig:image5}). This temporal granularity allows for analysis of both long-term climate trends and short-term weather patterns.

{\bf Population Density.} Leveraging the Gridded Population of the World, Version 4 (GPWv4) population dataset \citep{ciesin_gpwv4_2018}, which integrates censuses, population registers, and spatial distributions, we synthesize population density data for each point in the grid (\cref{fig:image7}).
\\

We emphasize \textsc{FAIR-Earth} is a \textit{unified} framework. All data is sampled and projected onto a uniform 0.1$^\circ$ x 0.1$^\circ$ grid, a state-of-the-art resolution for fine-grained analysis. Moreover, we propose attributes and metadata including landmass size, coast distance, and population density for each location, enabling us to disentangle the global prediction performance into meaningful subgroups (data specifications available in \cref{APP:MetadataDetails} and \cref{APP:FAIR-Earth Details}). In particular, our metadata comprises two main subgroups. First, {\em geographical features} are composed: based on a combination of existing data and ad-hoc definitions (\cref{APP:MetadataDetails}), we label relevant subgroups including islands, coastlines (stratified by distance), land, and sea. Additionally, we incorporate metadata on {\em population density} and {\em administrative boundaries}. 




While core geospatial data is derived from existing datasets, we highlight the differences between \textsc{FAIR-Earth} and existing stand-alone datasets (\cref{tab:dataset_comparison}). Unlike datasets such as NaturalEarth \citep{NaturalEarth} that emphasize geographic features, we combine multiple dimensions of data (environmental, demographic, and emissions) within one integrated format, allowing practitioners to model and assess the interdependence of multiple modalities. In contrast to existing climate datasets such as ERA5 \citep{hersbach1999era5}, our environmental signals incorporate interconnected emissions data and metadata, crucial for nuanced investigation of environmental justice issues.


\subsection{\textsc{FAIR-Earth Limitations}} While the design choices for \textsc{FAIR-Earth} facilitate subgroup-level fairness evaluations, they also yield certain tradeoffs. In particular, temporal and polar biases may manifest from the gridded format of \textsc{FAIR-Earth} data; we later discover (\ref{sec:tradeoffs}) that this format indeed induces bias in the training data. Given the agile framework of \textsc{FAIR-Earth}, we strongly encourage practitioners to evaluate and incorporate their own datasets when necessary, and to adopt best practices in sampling to mitigate such biases.


\begin{table}[t!]
\setlength{\tabcolsep}{4pt}
\label{tab:test_loss_table}
\caption{Average subgroup cross-entropy test loss of \textsc{Spherical Harmonic} across cross-validation, stratified by spatial resolution. {\bf We observe that the bias of the model in missing ``island'' and ``coastline'' persists even for high resolution dataset, as improvement plateaus.} }
\begin{center}
\begin{sc}
\small
\begin{tabular}{l||c|c|c|c|c|c}
\toprule
Resolution & 5000 & 10000 & 15000 & 20000 & 25000 & 30000 \\
\midrule
Total & 0.16 $\pm$ 0.02 & 0.13 $\pm$ 0.03 & 0.12 $\pm$ 0.03 & 0.11 $\pm$ 0.04 & 0.10 $\pm$ 0.04 & 0.10 $\pm$ 0.04 \\
Land & 0.21 $\pm$ 0.03 & 0.15 $\pm$ 0.04 & 0.14 $\pm$ 0.05 & 0.14 $\pm$ 0.05 & 0.12 $\pm$ 0.06 & 0.16 $\pm$ 0.06 \\
Sea & 0.11 $\pm$ 0.02 & 0.16 $\pm$ 0.02 & 0.09 $\pm$ 0.03 & 0.08 $\pm$ 0.03 & 0.08 $\pm$ 0.03 & 0.80 $\pm$ 0.03 \\
Island & 2.74 $\pm$ 0.03 & 3.25 $\pm$ 0.49 & 2.85 $\pm$ 0.33 & 2.66 $\pm$ 0.26 & 2.61 $\pm$ 0.20 & 2.51 $\pm$ 0.21 \\
Coastline & 1.06 $\pm$ 0.08 & 1.00 $\pm$ 0.06 & 0.95 $\pm$ 0.05 & 0.90 $\pm$ 0.04 & 0.82 $\pm$ 0.04 & 0.81 $\pm$ 0.05 \\
\bottomrule
\end{tabular}
\end{sc}
\end{center}
\end{table}

\section{Measuring and Improving the (Un)Fairness of Implicit Neural Representations}

\subsection{Evaluation of Existing Representations}
\label{Evaluation}

\paragraph{Global Performance is not Representative of Local Performance}
\label{evaluation-notrepresentative}
To properly understand the behavior of existing INRs, we depart from hyperparameter fine-tuning, and instead perform extensive training over a cross-product of each model's hyperparameter space. This allows us to move past ``best-case" analysis, and instead incorporate nuanced analysis of model tradeoffs. Moreover, as downstream tasks often require further tuning regardless, it is natural to instead consider the overall behavior of the implicit representation. 
Experimental details are available in \cref{APP:Training Specifications}.

In particular, we explore the correlations of subgroup performance. While certain disparities in performance are intrinsic to the task (e.g., islands and coastlines exhibit higher representation loss as land-sea boundaries are harder to learn than constant signals), a \textit{fair} model should in general show \textit{concurrent improvement in subgroups}.
Leveraging the metadata in \textsc{FAIR-Earth}, we compare representation performance between areas that correspond to local signals (e.g., islands) and ones that correspond to global signals (e.g., large landmasses).

\begin{table}[t!]
    \centering
    \label{tab:pearson_comparison}
    \caption{Land-island performance correlations across encodings and training resolutions. While existing encodings compromise local-global performance, \textsc{Spherical Wavelet} reconciles improvement over multi-scale features. We omit \textsc{SphereC+} due to abnormally high bias.}
    \begin{tabular}{
        l
        S[table-format=-1.2]
        S[table-format=-1.2]
        S[table-format=-1.2]
    }
    \toprule
    \multirow{2}{*}{Training Samples} & 
    \multicolumn{3}{c}{\textbf{Pearson Correlation (R)}} \\
    \cmidrule(l){2-4}
    & {\textsc{SW} (Ours)} & {\textsc{SH}} & {\textsc{Theory}} \\
    \midrule
    5000   & 0.51 & -0.73 & -0.53 \\
    10000  & 0.42 &  -0.88 & -0.62 \\
    15000  & 0.04 &  -0.88 & -0.10 \\
    20000  &  0.31 &  -0.68 &  -0.28 \\
    \bottomrule
    \end{tabular}
\end{table}

Strikingly, our correlation analysis indicates a stark disparity in representation loss between global and local groups among \textit{all state-of-the-art encodings} (\cref{tab:test_loss_table}, \cref{tab:pearson_comparison}).
For \textsc{Spherical Harmonics}, we notice a strong negative correlation between land and island loss when stratified along training resolution (\cref{tab:pearson_comparison}),  suggesting that when optimized for total loss, state-of-the art INRs exhibit a clear tradeoff between global and local performance.
Regardless of stratification, INRs of the form \cref{eq:decomposition} appear incapable of \textit{competitive} performance across all subgroups at once (\cref{tab:spherical-comparison}).

\begin{figure}[t!]
    \centering
    \begin{tabular}{cc}
        \includegraphics[width=0.48\textwidth]{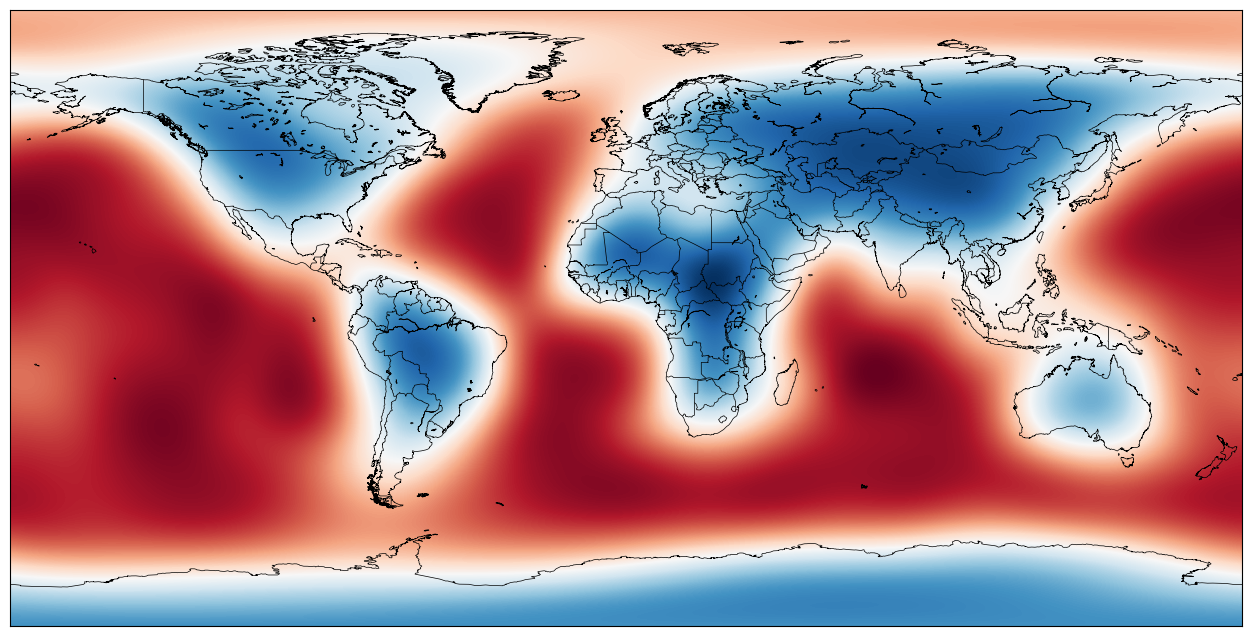} &
        \includegraphics[width=0.48\textwidth]{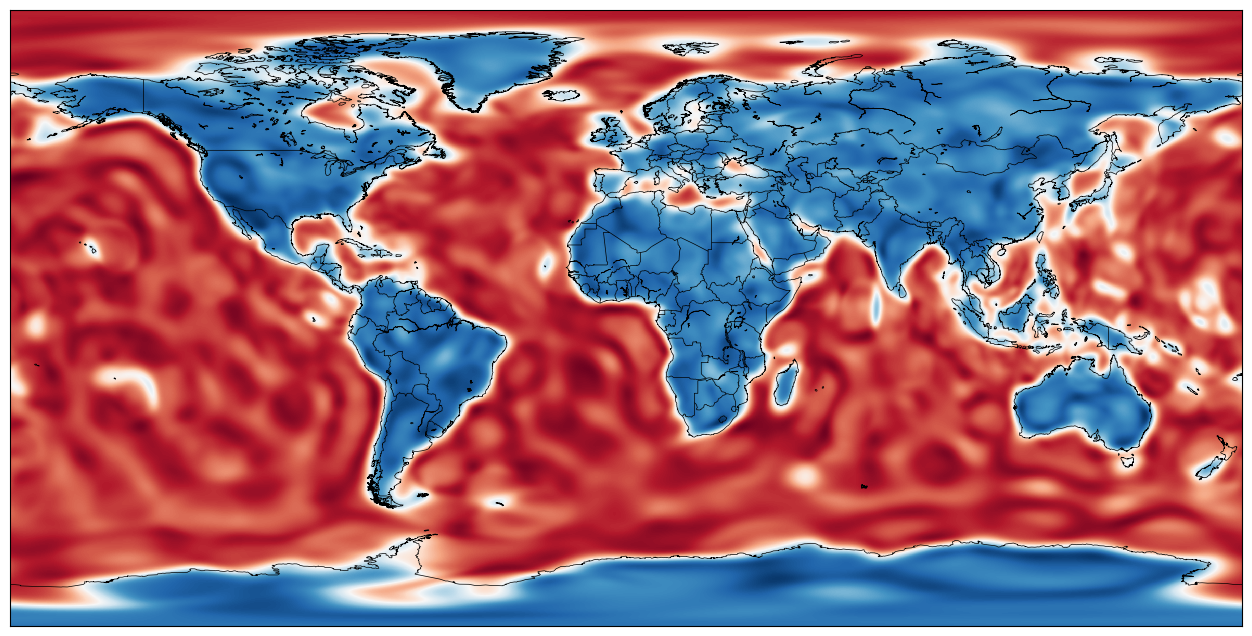} \\
        a) High regularization \textit{implies} low resolution &
        b) Low regularization \textit{implies} aliasing \\
    \end{tabular}
    \caption{Model behavior at different resolutions and regularizations. Smaller models fail to capture fine/local signals. Larger models poorly reconcile local signals with existing global signals.}
    \label{fig:sidebyside}
\end{figure}

\vspace{-0.2cm}
\paragraph{Aliasing as a Result of Overfitting}
\label{evaluation-aliasing}
\vspace{-0.2cm}

As described earlier, \cref{eq:decomposition} is global in nature. We observe the implications of this construction in \cref{fig:sidebyside}, where the smoothness of land and sea signals are compromised in an effort to represent islands and coastlines. 
We corroborate this observation with an ablation over encoding size, which shows increasing encoding size results in plateauing land performance and deteriorating island performance (\cref{fig:sizeAblation}).

Concretely, the observed aliasing in the high-resolution case is indicative of the Gibbs phenomenon, exhibiting unnatural oscillations around the island and coastline `discontinuities' \citep{dyke2001introduction}. In the effort to precisely fit the islands, \cref{eq:decomposition} is forced to learn an \textit{unnatural} representation of the land and sea masses. 

Our correlation analysis and aliasing investigation confirms that no \textsc{Spherical Harmonic}-based model is able to maintain competitive performance in both sub-groups simultaneously, suggesting that current solutions in implicit Earth representations require further development to reach truly equitable predictions.

\begin{figure}[t!]
    \centering
    \begin{tabular}{cc}
        \includegraphics[width=0.48\textwidth]{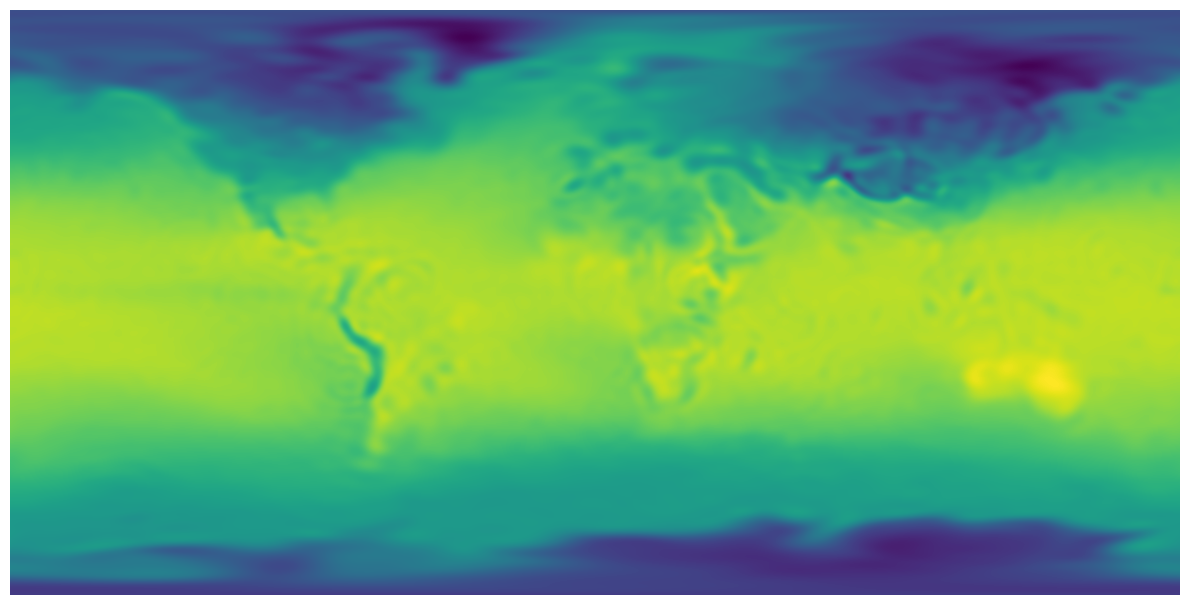} &
        \includegraphics[width=0.48\textwidth]{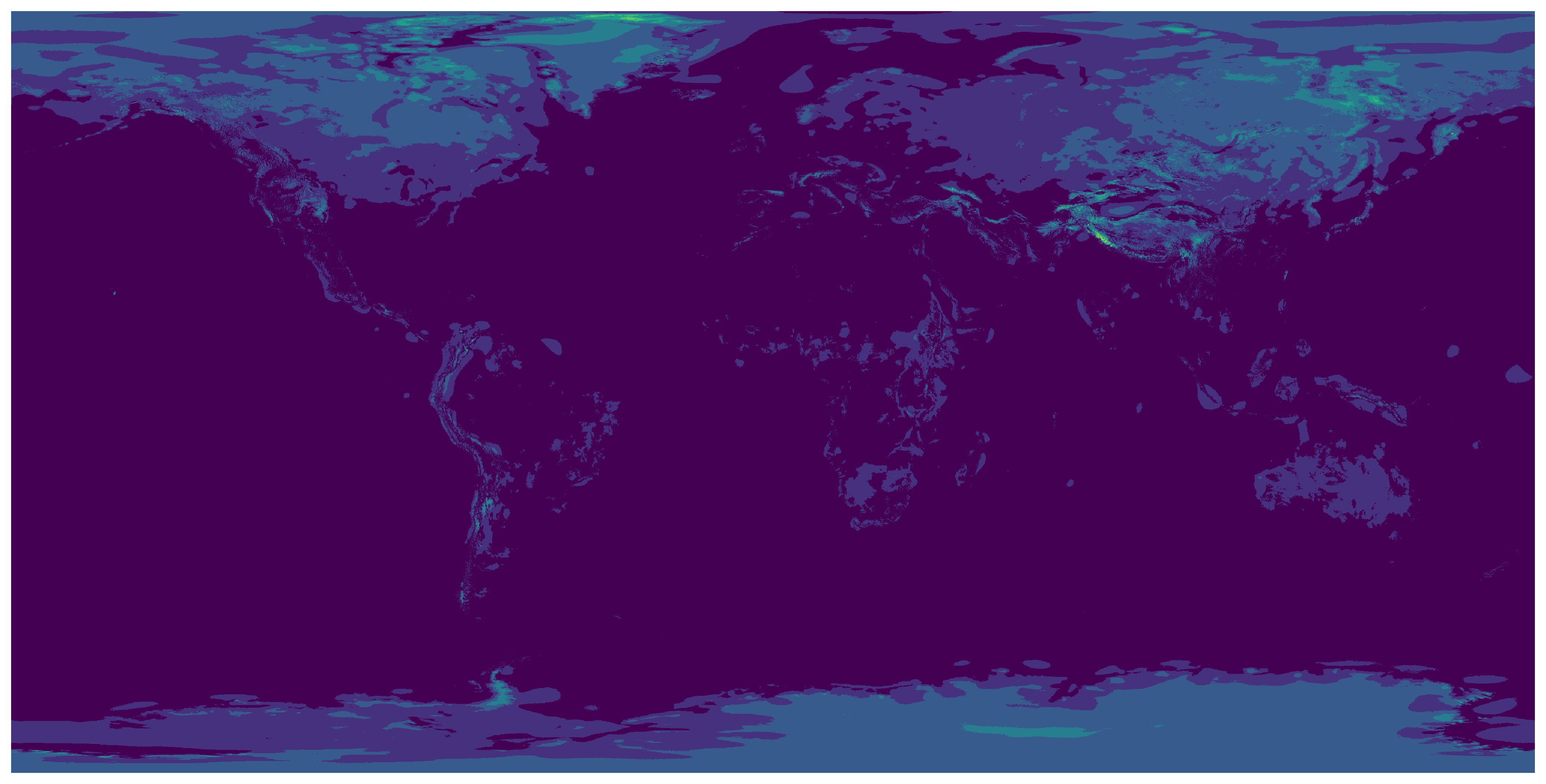} \\
        a) Predictions Results &
        b) (Binned) Error Plot \\
    \end{tabular}
    \caption{Qualitative performance of \textsc{Spherical Harmonic} on surface temperature regression task. \textsc{SH} exhibit bias against high-frequency details, with loss concentrated in areas of abrupt change (e.g., coastlines).}
    \label{fig:highFreqTAS}
\end{figure}

\paragraph{Biases Across Multiple Modalities and Subgroups}
\label{evaluation-multiplemodalities}
While the land-sea binary classification task reveals natural biases against fine and localized areas, we emphasize that one of \textsc{FAIR-Earth}'s main strengths is in its ability to easily quantify subgroup disparities across multiple modalities.
In particular, we perform similar stratified evaluation against \textsc{FAIR-Earth}'s benchmarks for environmental signals. For the surface temperature dataset, which exhibits similar sharp variations across recognizable boundaries, we note similar trends (\cref{fig:highFreqTAS}). Namely, \textsc{FAIR-Earth} reveals systematic patterns in representation quality: regions with sharper variations, particularly near the coast, show significantly higher average representation loss ($\text{MSE}_\text{Land}=0.87$, $\text{MSE}_\text{Coast}=0.101$) compared to regions with smoother variations ($\text{MSE}_\text{Sea}=0.43$, $\text{MSE}_\text{Island}=0.49$) (\cref{tab:subgroup-temperature}).

Moreover, this analysis extends naturally to downstream biases through \textsc{FAIR-Earth}'s rich metadata. At the country level, we observe that representation challenges at the feature level manifest as systematic performance disparities. For instance, \textsc{Spherical Harmonic} and \textsc{Theory} encodings particularly struggle with to demarcate the Spain's fine Mediterranean coastline, while all studied encodings show degraded performance in coastal countries due to sharp temperature gradients at land-sea boundaries (\ref{tab:encoding-comparison}).

\subsection{Spherical Wavelets for Fair Earth Representations}
\label{SW}

As explored in \cref{Background}, a major limitation of \textsc{Spherical Harmonic} was their global support, which struggled to reconcile localized signals with global ones. In particular, \textsc{SH} required a large amount of basis functions to properly represent localized signals, and this excessive parameterization resulted in aliasing in other regions. The ad-hoc solution involving re-weighting subgroups depends on extensive labels; moreover, the effect on global performance is unknown \citep{stone2024epistemicuncertaintyweightedlossvisual}.

To this end, we are motivated to leverage the theoretical guarantees of wavelets, which allow for multi-scale resolution analysis of signals, efficiently representing signals at various scales and locations. To extend towards Earth data, we refer to past work in \citet{2003SPIE.5207..208D} that introduces spherical wavelet basis functions. Similar to the formulation in \cref{eq:decomposition}, any function on the sphere can also be approximated by a discrete sum of wavelet basis functions, where coefficients are similarly learned via the INR mechanism. 

The following sections briefly outline the construction of a novel encoding mechanism \textsc{Spherical Wavelet}, and its seamless integration into the currently studied INR pipeline. We direct interested readers towards \citet{McEwen_2007, 2003SPIE.5207..208D} for analysis on the correctness of the construction, and towards \ref{sec:wavelet-notes} for evidence that \textsc{SW} encodings are surprisingly more \textit{efficient} and \textit{stable} than their \textsc{SH} counterparts.

{\bf Spherical Morlet Wavelet via Inverse Projection}~
\label{sw-projection}
A concise, computationally tractable construction for spherical wavelets arises from inverse stereographic projection of Euclidean wavelets \citep{2003SPIE.5207..208D}. We note that due to the nature of this projection, the \textsc{Spherical Wavelet} is ill-defined at the poles.

As proven in \citet{sanz2006waveletssphereapplicationdetection}, this inverse stereographic projection $\Pi^{-1}$ of an admissible \textit{Euclidean} wavelet yields an admissible \textit{spherical} wavelet; that is, the projected wavelet $\psi(\theta,\varphi)$ satisfies the necessary zero-mean condition \citep{2003SPIE.5207..208D}.
Empirically, we find that the Morlet mother wavelet shows consistently superior performance (\cref{fig:wavelet-filter-perf}). Thus, applying the inverse projection onto the 2D Morlet wavelet with width factor and wave number $w, k=1$ yields the admissible spherical Morlet mother wavelet: 
\begin{align}
\psi_M(\theta, \phi) = [\Pi^{-1}\psi_{\mathbb{R}^2}](\theta, \phi) =  \frac{e^{i\tan(\theta/2)\cos(\phi)} e^{-(1/2)\tan^2(\theta/2)}}{1 + \cos\theta} \label{eq:motherWavelet}
\end{align}


\begin{figure}[t!]
    \centering
    \begin{tabular}{cc}
        \includegraphics[width=0.3\textwidth]{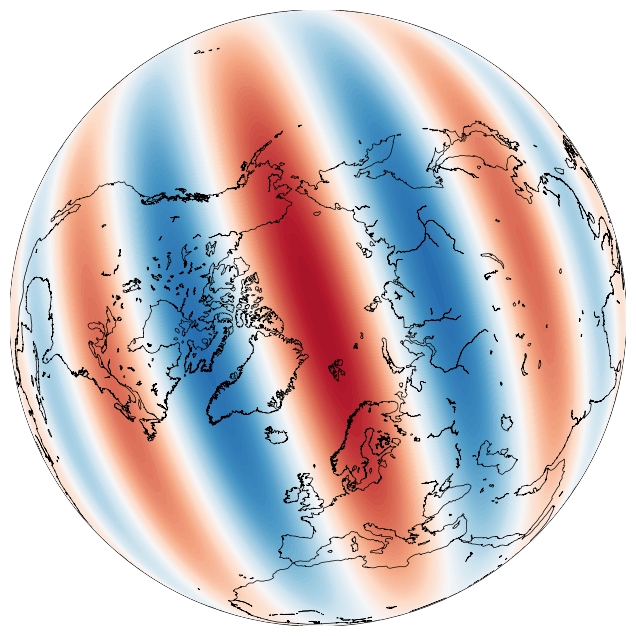} &
        \includegraphics[width=0.3\textwidth]{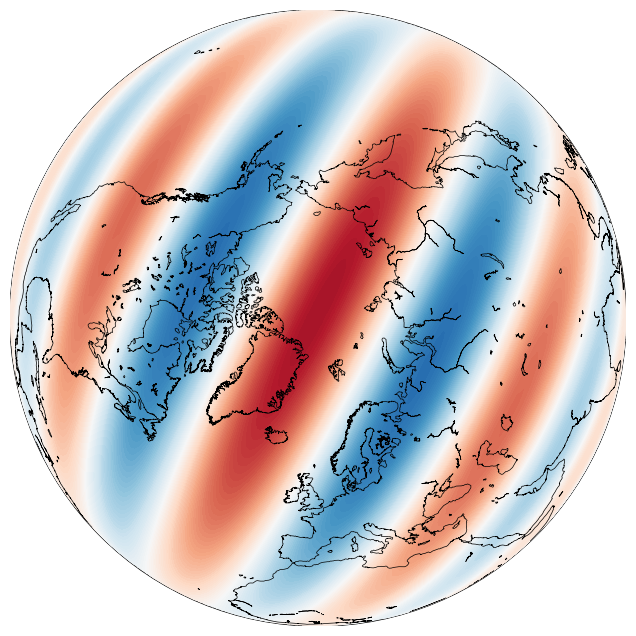} \\
        a) Mother Wavelet (\cref{eq:motherWavelet}) &
        b) Rotated Wavelet \\[1ex]
        \includegraphics[width=0.3\textwidth]{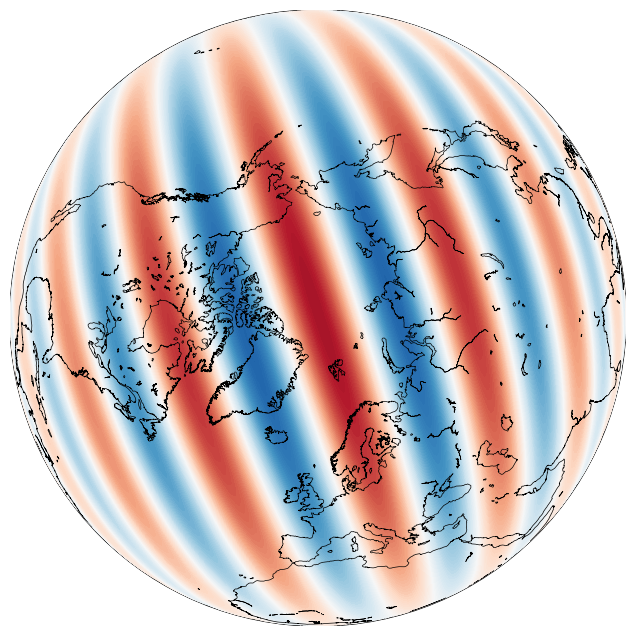} &
        \includegraphics[width=0.3\textwidth]{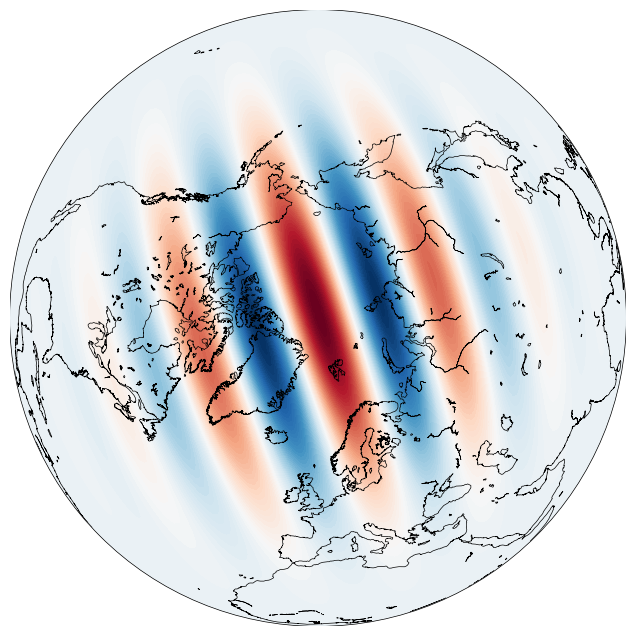} \\
        c) Wave number controls central \\ frequency of the Gaussian envelope &
        d) Dilation factor controls energy concentration
    \end{tabular}
    \caption{\textsc{Spherical Wavelet} parameterization. All filters are identical to (\cref{eq:motherWavelet}) except for one change in the specified parameter.}
    \label{fig:SWbanks}
\end{figure}

{\bf Lifting Scheme}~
\label{SW-lifting}
From an admissible mother wavelet, we can now construct a wavelet basis from affine transformation on the sphere. 
Analogous to time-frequency localization of Euclidean wavelets, spherical wavelets provide rotation-dilation localization. The rotation operator $\mathcal{R}(\rho) \equiv \mathcal{R}(\alpha, \beta, \gamma)$ and dilation operator $\mathcal{D}(a)$ both transport a function from and into $f \in L^2(\mathbb{S}^2, d\mu(\theta, \phi))$; refer to \citet{McEwen_2007} for rigid formulation. 
Applying these transformations, we then define a set of orthogonal basis functions
$\{\psi_{a,\rho} \equiv \mathcal{R}(\rho)\mathcal{D}(a)\psi_M\}$

Finally, we discretize the reconstruction formula. First, we deterministically distribute $\rho$ across $N$ points on the sphere following the Fibonacci lattice $\mathcal{F}_N$ \citep{Gonz_lez_2009}. To discretize rotation, we take $a \in \{2^{\frac{i}{Q}}, 0 < i \leq M \}$, where $Q \leq 8$ is a fixed value \citep{mallat1999wavelet}. This yields an embedding size on the order of $\mathcal{O}(NM)$, and the discretized approximation
$$f(\theta, \phi)_{N, M} \equiv \sum_{\rho \in \mathcal{F}_N} \sum_{i=1}^{M} w_{a, \rho} \psi_{a, \rho}(\theta, \phi)$$

{\bf Localization Properties}~
\label{SW-localization}
The spherical Morlet wavelet, by design, exhibits dual localization in both dilation and rotation, offering significant advantages in signal representation on the sphere \citep{McEwen_2007}. These properties enable a more nuanced and adaptive analysis of spherical data. In particular, the Morlet variant is parameterized as follows (\cref{fig:SWbanks}):
\begin{itemize}
    \item Dilation allows the wavelet to adapt its scale, effectively increasing the resolvable frequency range of our representation.
    \item Rotation facilitates representation of directional and localized features on the sphere.
    \item The wave number and scale factors parameterize oscillation.
\end{itemize}

{\bf Computational Comparison}~
While wavelets are more heavily parameterized than their Fourier counterparts, one might expect them to be more computationally expensive to compute and tune. However, empirical evidence shows that hyperparameter searches converge at similar rates (\cref{fig:fine-tuning-dynamics}), and \textsc{Spherical Wavelet} demonstrates superior efficiency for large encodings (\cref{tab:encoding_times}). Refer to \cref{sec:wavelet-notes} for an in-depth analysis.

\begin{table}[t!]
   \centering
   \caption{Computational time for \textsc{Spherical Harmonic} and \textsc{Spherical Wavelet} across different sizes, with faster times underlined. \textsc{Spherical Wavelet} scales better than \textsc{Spherical Harmonic}, which can be attributed to its simpler formulation.}
   \begin{tabular}{c|cc|ccc}
       \toprule
       & \multicolumn{2}{c|}{Generation Time (ms)} & \multicolumn{3}{c}{Encoding Parameters} \\
       Size & \textsc{SH} & \textsc{SW} & Legendre Polyn. & Dilations & Rotations \\
       \midrule
       25 & \underline{2.60 $\pm$ 0.08} & 5.83 $\pm$ 0.67 & 5 & 1 & 25 \\
       100 & \underline{16.44 $\pm$ 0.72} & 22.67 $\pm$ 2.85 & 10 & 4 & 25 \\
       625 & 220.13 $\pm$ 9.05 & \underline{129.58 $\pm$ 18.43} & 25 & 5 & 125 \\
       900 & 368.73 $\pm$ 17.93 & \underline{204.25 $\pm$ 27.05} & 30 & 6 & 150 \\
       \bottomrule
   \end{tabular}
   \label{tab:encoding_times}
\end{table}

\subsection{Spherical Wavelet Experiments}
\paragraph{\textsc{Spherical Wavelet} Corrects Localized Biases}
\label{sw-biases}

To examine the biases present, we begin by analyzing the trends in training and evaluation.
In particular, we leverage \textsc{FAIR-Earth} and deploy a large cross-product of parameters ($N=984$) to cross-validate algorithm performance between local signals and global signals over various configurations. As a baseline, we compare to results to $[\textsc{Spherical Harmonic}, \textsc{Theory}, \text{ and } \textsc{Grid and Sphere}]$ encodings trained under identical settings, therefore isolating the effects of the encodings.

We observe a sharp contrast between the local-global correlation of our encodings; that is, \textsc{Spherical Wavelet} learns to \textit{efficiently} represent islands even at coarser resolution, therefore inducing a positive trend between global (land) and localized (island) performance, whereas our baseline encodings exhibit a tradeoff between global/local performance even for finer resolutions \cref{tab:pearson_comparison}.

\begin{figure}[t!]
    \centering
    \begin{tabular}{cc}
        \includegraphics[width=0.3\linewidth]{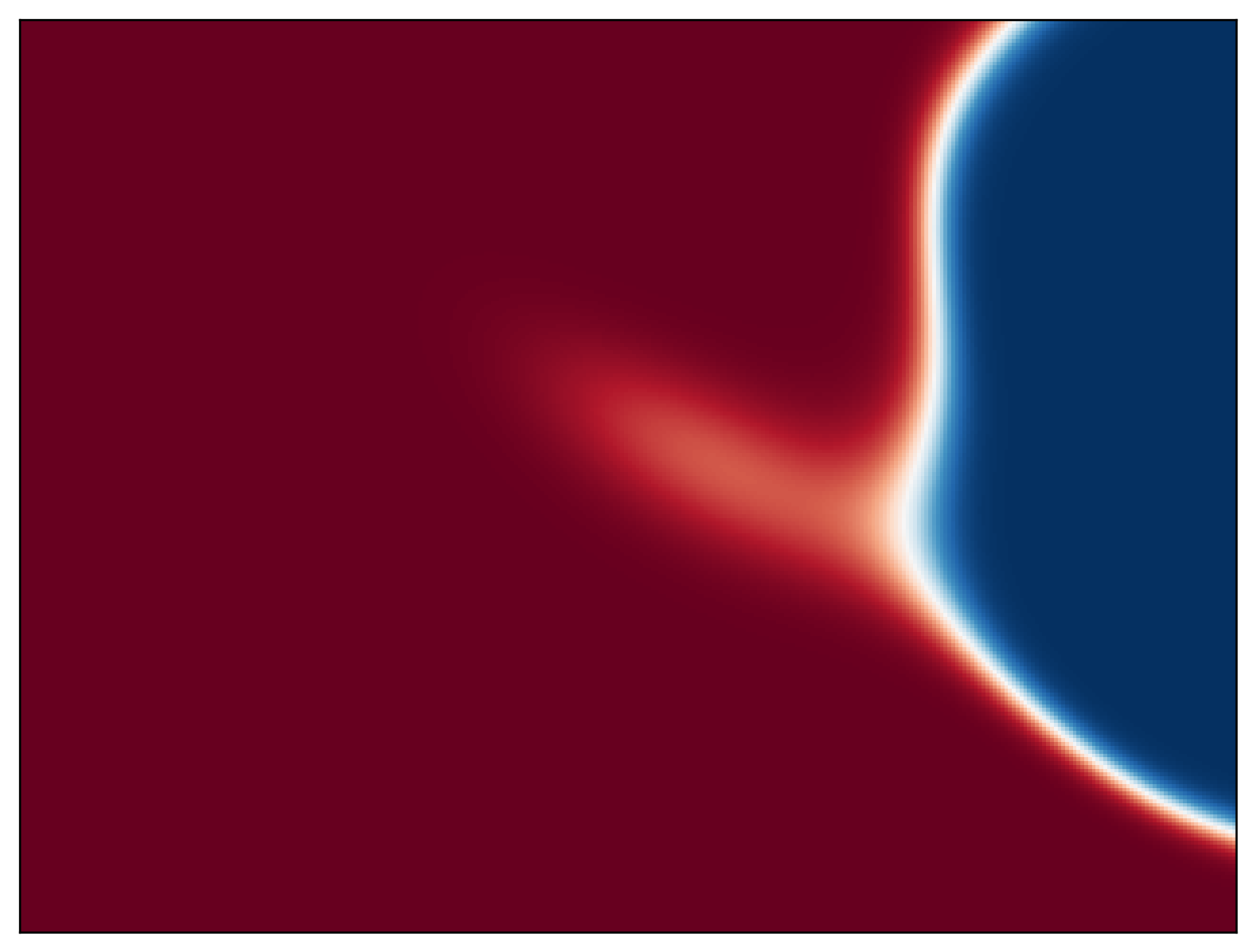} &
        \includegraphics[width=0.3\linewidth]{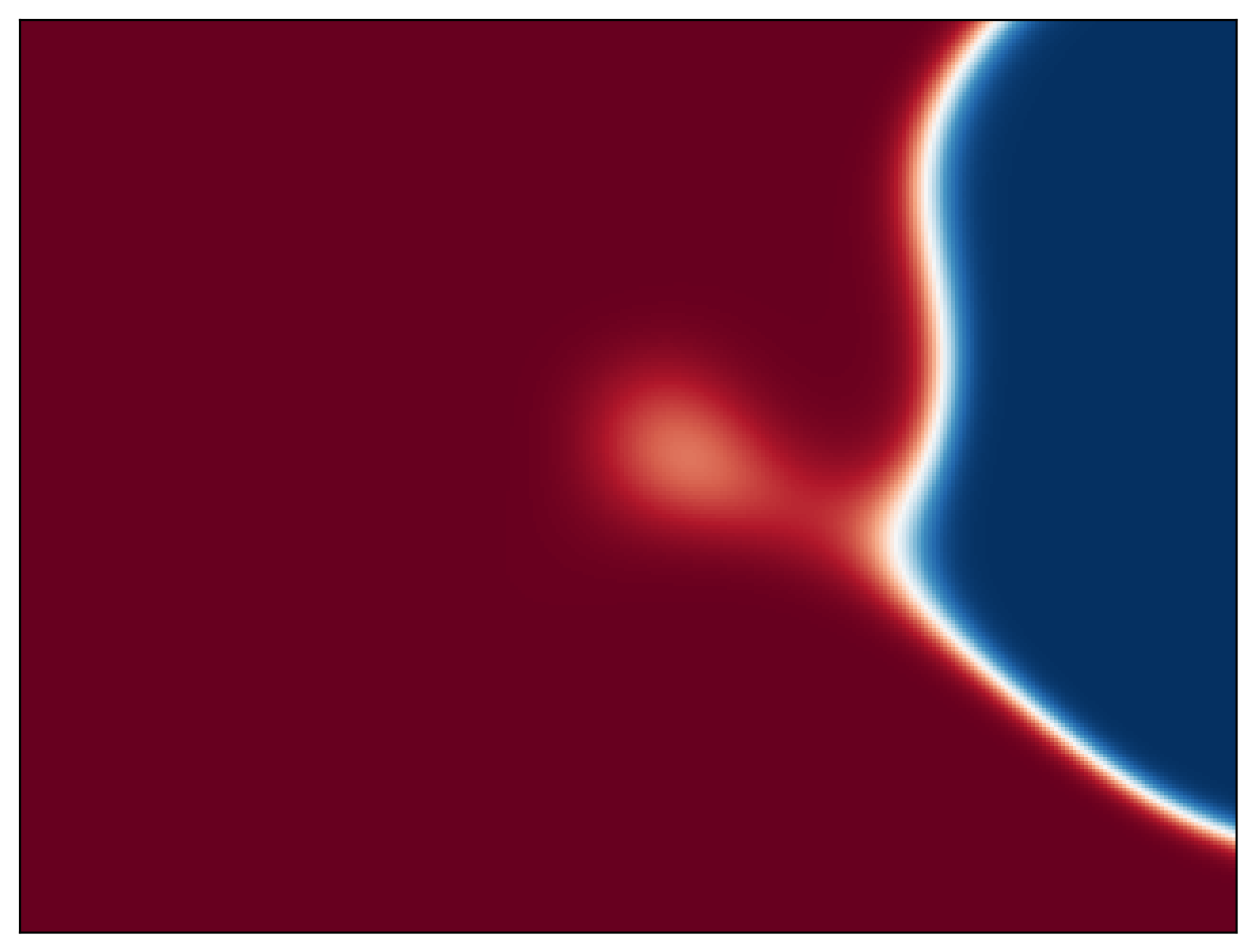} \\
        \includegraphics[width=0.3\linewidth]{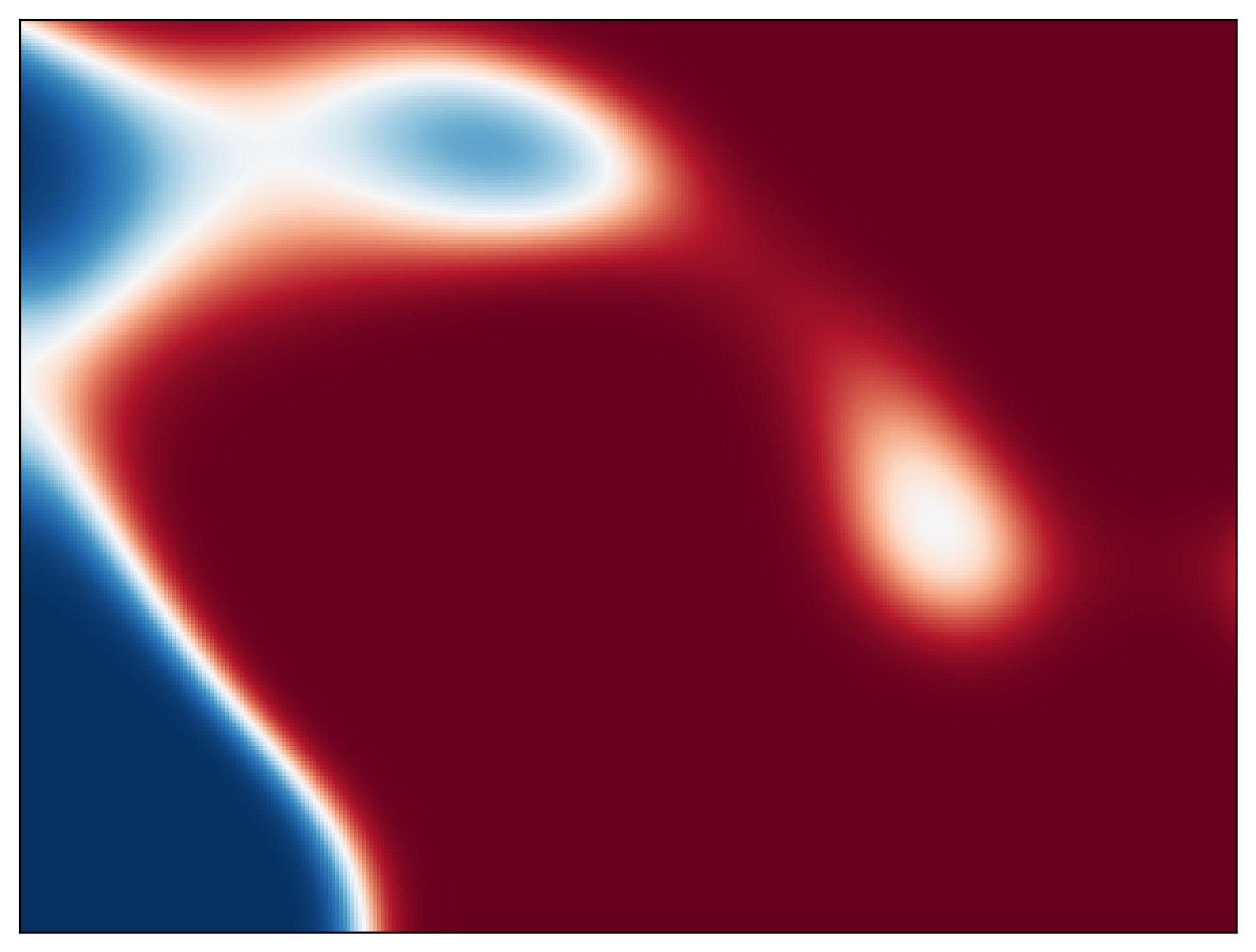} &
        \includegraphics[width=0.3\linewidth]{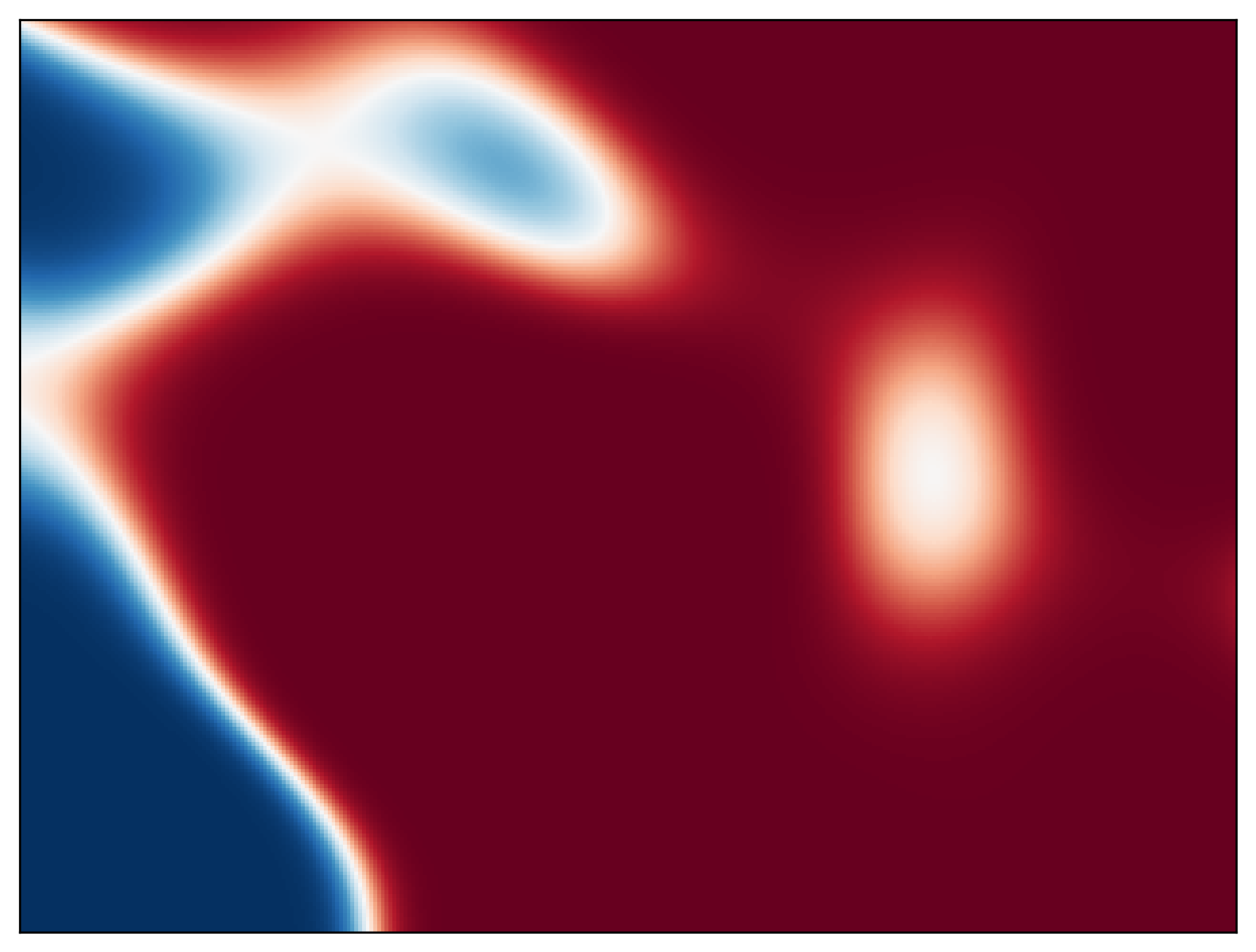} \\
        \includegraphics[width=0.3\linewidth]{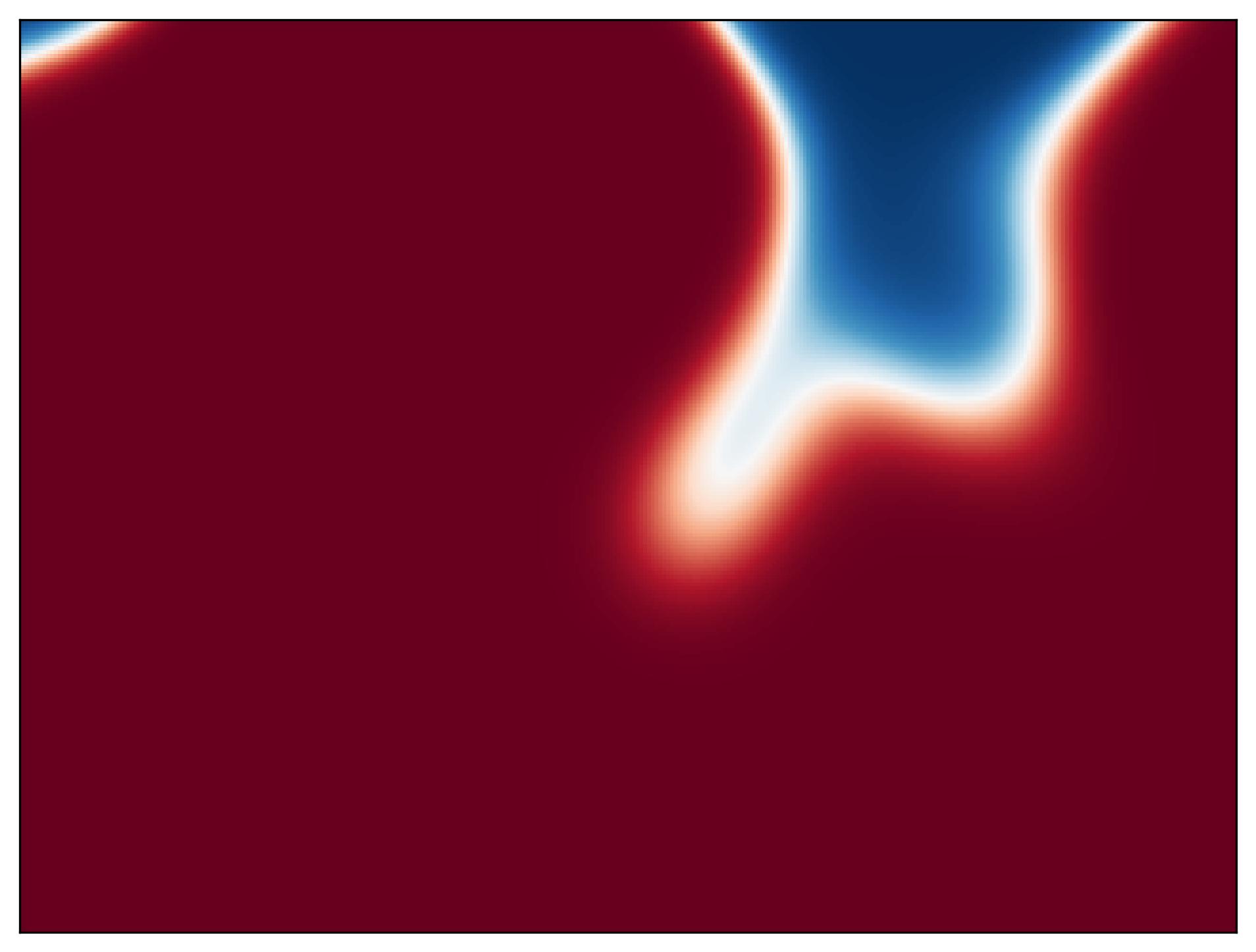} &
        \includegraphics[width=0.3\linewidth]{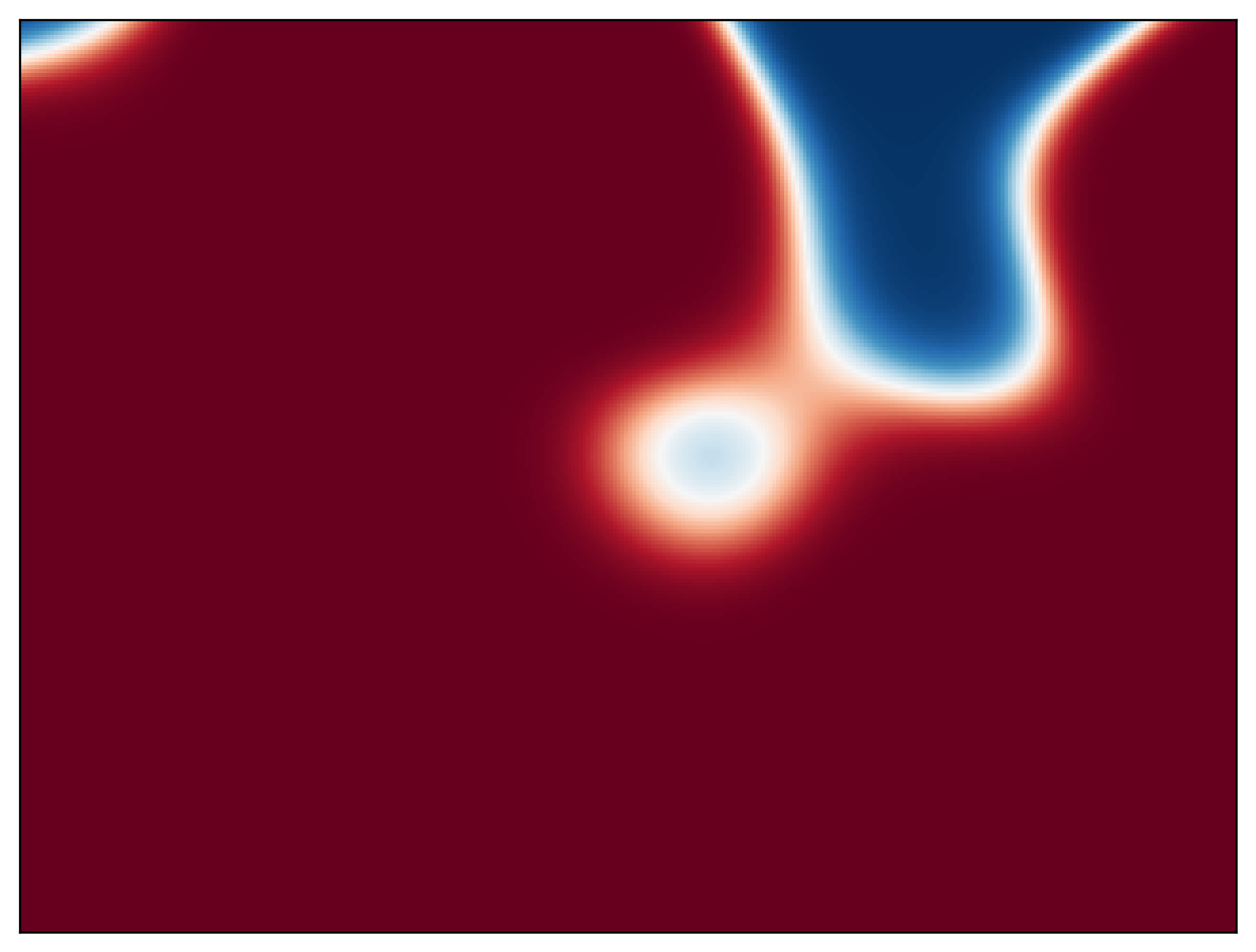} \\[1ex]
        a) \textsc{Spherical Harmonic} &
        b) \textsc{Spherical Wavelet}
    \end{tabular}
    \caption{Comparisons of \textsc{Spherical Harmonic} and \textsc{Spherical Wavelet} behavior in localized regions. Full figures available in ~\cref{fig:SHprediction} and~\cref{fig:SWprediction}.}
    \label{fig:visual_comparison}
\end{figure}

Specifically, in contrast to \textsc{Spherical Harmonic}, where finer features are often blended together (\cref{fig:failure_cases}), \textsc{Spherical Wavelet} has a significantly higher resolvable resolution for similar parameter counts. Especially in localized areas such as islands, our encodings tend to produce sharper, more fine boundaries compared to spherical harmonic encodings (\cref{fig:visual_comparison}).

\paragraph{\textsc{Spherical Wavelet} is Competitive with Existing Methods}~ \label{SW-competitive} While our correlation analysis showed performance of \textsc{Spherical Wavelet} exhibited fair and stable \textit{trends}, we also seek to show performance is competitive with existing methods. First, we re-examine the previous experiment and compare general global performance of \textsc{Spherical Wavelet} and \textsc{Spherical Harmonic} across well-defined hyperparameter spaces. 

In general, we observe that the learned representation is similar to that of spherical harmonics, as indicated by the visual similarities in their learned representations (\cref{fig:visual_comparison}). 
However, \textsc{SW} in fact exhibits \textit{competitive or superior} performance over \textsc{SH} through various training resolutions: we maintain significantly superior average performance across all training resolutions (\cref{tab:spherical-comparison}). 

\vspace{0.5em}
\begin{table}[t!]
\caption{\centering Comparison of best/average cross-entropy loss between \textsc{SH} and \textsc{SW} over land-sea cross-validation. Top encodings per stratification are bolded.}
\centering
\renewcommand{\arraystretch}{1.5}
\setlength{\tabcolsep}{12pt}
\begin{tabular}{l c c}
\toprule
\textbf{Samples} & \textbf{\textsc{Spherical Wavelet}} & \textbf{\textsc{Spherical Harmonic}} \\
\midrule
5000  & 0.1404 / \textbf{0.1473} & \textbf{0.1392} / 0.1524 \\
10000 & \textbf{0.1111} / \textbf{0.1169} & 0.1137 / 0.1218 \\
15000 & \textbf{0.1000} / \textbf{0.1056} & 0.1003 / 0.1086 \\
20000 & 0.0912 / \textbf{0.0962} & \textbf{0.0892} / 0.0969 \\
\bottomrule
\end{tabular}
\vspace{0.5em}
\label{tab:spherical-comparison}
\end{table}

The degradation in performance for \textsc{Spherical Wavelet} for higher resolutions may be attributed to the misspecifications of certain parameters in our model (e.g. wave number, scaling factors).

To this end, we also evaluate fine-tuned performance on the temperature and precipitation regression tasks, observing more consistent performance: when properly parameterized, \textsc{Spherical Wavelet} performs roughly on par with \textsc{Spherical Harmonic} encodings across all \textsc{FAIR-Earth} tasks (\cref{sec:Experimental-Results}).

\paragraph{\textsc{Spherical Wavelet} Tradeoffs} 
\label{sec:tradeoffs}
While \textsc{Spherical Wavelet} demonstrates robust performance across various tasks, like all signal-based methods, it faces certain limitations. We investigate these limitations using two challenging external datasets:

\begin{itemize}
    \item On the checkerboard classification task \cite{rußwurm2024geographiclocationencodingspherical}, which features coarse, single-scale signals, \textsc{Spherical Wavelet} performs well relative to most baselines but consistently falls short of \textsc{Spherical Harmonic}. This suggests that the wavelet decomposition may introduce unnecessary complexity for simple, single-scale tasks.
    
    \item In the land-sea classification task \cite{rußwurm2024geographiclocationencodingspherical}, which uses non-gridded, ultra high-resolution data, \textsc{Spherical Wavelet} only marginally outperforms \textsc{Spherical Harmonic}. Our latitudinal analysis (\cref{fig:latitudeAblation}) reveals that this limited improvement stems primarily from degraded performance near the poles.
\end{itemize}

\section{Conclusion}
Measuring the fairness of implicit Earth representations is pressing yet lacking field in both the Earth science and implicit representation fields. Our contributions in measuring and improving certain biases in state-of-the-art INRs represents a step forward in this important direction. By introducing \textsc{FAIR-Earth}, a novel framework for high-resolution assessment of subgroup-level performance, we aim to develop and assess models pertinent to the needs of practitioners, informing of existing biases and limitations of Earth models.
Leveraging this new framework, our comprehensive assessment uncovers striking biases against certain subgroups across state-of-the-art INRs.  
Based on these observed performance disparities, we propose \textsc{Spherical Wavelet}, injecting localized location encodings that are naturally motivated. Against the many benchmarks of \textsc{FAIR-Earth}, we resolve many of the biases evident in \textsc{Spherical Harmonic}, while maintaining competitive performance.
We invite iteration and updates on \textsc{FAIR-Earth} to improve the coverage of subgroups and signals, and stay relevant with changes and trends within the community. Additionally, given the limitations of \textsc{Spherical Wavelet}, we hope further research can develop heuristics or structural changes to address these shortcomings; in particular, we point to the integration of non-projection based \textsc{Spherical Wavelet} constructions as a promising direction for future work \cite{sanz2006waveletssphereapplicationdetection}.

\bibliography{iclr2025_conference}

\begin{thebibliography}{48}
\providecommand{\natexlab}[1]{#1}
\providecommand{\url}[1]{\texttt{#1}}
\expandafter\ifx\csname urlstyle\endcsname\relax
  \providecommand{\doi}[1]{doi: #1}\else
  \providecommand{\doi}{doi: \begingroup \urlstyle{rm}\Url}\fi

\bibitem[Nat()]{NaturalEarth}
Natural earth.
\newblock \url{https://www.naturalearthdata.com/}.

\bibitem[Allen et~al.(2002)Allen, Kettleborough, and Stainforth]{allen2002model}
Myles~R Allen, JA~Kettleborough, and DA~Stainforth.
\newblock Model error in weather and climate forecasting.
\newblock In \emph{ECMWF Predictability of Weather and Climate Seminar}, pp.\  279--304. European Centre for Medium Range Weather Forecasts, Reading, UK, 2002.

\bibitem[Balestriero et~al.(2022)Balestriero, Bottou, and LeCun]{balestriero2022effectsregularizationdataaugmentation}
Randall Balestriero, Leon Bottou, and Yann LeCun.
\newblock The effects of regularization and data augmentation are class dependent, 2022.
\newblock URL \url{https://arxiv.org/abs/2204.03632}.

\bibitem[Beck et~al.(2019)Beck, Wood, Pan, Fisher, Miralles, van Dijk, McVicar, and Adler]{beck2019mswep}
H.~E. Beck, E.~F. Wood, M.~Pan, C.~K. Fisher, D.~G. Miralles, A.~I. J.~M. van Dijk, T.~R. McVicar, and R.~F. Adler.
\newblock Mswep v2 global 3-hourly 0.1° precipitation: Methodology and quantitative assessment.
\newblock \emph{Bulletin of the American Meteorological Society}, 100:\penalty0 473--500, 2019.
\newblock \doi{10.1175/BAMS-D-17-0138.1}.

\bibitem[{Center for International Earth Science Information Network - CIESIN - Columbia University}(2018{\natexlab{a}})]{ciesin2018gpw}
{Center for International Earth Science Information Network - CIESIN - Columbia University}.
\newblock Gridded population of the world, version 4 (gpwv4): Population density adjusted to match 2015 revision un wpp country totals, revision 11.
\newblock Technical report, NASA Socioeconomic Data and Applications Center (SEDAC), Palisades, New York, 2018{\natexlab{a}}.
\newblock Accessed 29 August, 2024.

\bibitem[{Center for International Earth Science Information Network - CIESIN - Columbia University}(2018{\natexlab{b}})]{ciesin_gpwv4_2018}
{Center for International Earth Science Information Network - CIESIN - Columbia University}.
\newblock Gridded population of the world, version 4 (gpwv4): Population density adjusted to match 2015 revision un wpp country totals, revision 11, 2018{\natexlab{b}}.
\newblock URL \url{https://doi.org/10.7927/H4F47M65}.

\bibitem[Chen et~al.(2023)Chen, Wu, Grinspun, Zheng, and Chen]{chen2023implicitneuralspatialrepresentations}
Honglin Chen, Rundi Wu, Eitan Grinspun, Changxi Zheng, and Peter~Yichen Chen.
\newblock Implicit neural spatial representations for time-dependent pdes, 2023.
\newblock URL \url{https://arxiv.org/abs/2210.00124}.

\bibitem[Cole et~al.(2023)Cole, Van~Horn, Lange, Shepard, Leary, Perona, Loarie, and Mac~Aodha]{cole2023spatial}
Elijah Cole, Grant Van~Horn, Christian Lange, Alexander Shepard, Patrick Leary, Pietro Perona, Scott Loarie, and Oisin Mac~Aodha.
\newblock Spatial implicit neural representations for global-scale species mapping.
\newblock In \emph{International Conference on Machine Learning}, pp.\  6320--6342. PMLR, 2023.

\bibitem[{Demanet} \& {Vandergheynst}(2003){Demanet} and {Vandergheynst}]{2003SPIE.5207..208D}
Laurent {Demanet} and Pierre {Vandergheynst}.
\newblock {Gabor wavelets on the sphere}.
\newblock In Michael~A. {Unser}, Akram {Aldroubi}, and Andrew~F. {Laine} (eds.), \emph{Wavelets: Applications in Signal and Image Processing X}, volume 5207 of \emph{Society of Photo-Optical Instrumentation Engineers (SPIE) Conference Series}, pp.\  208--215, November 2003.
\newblock \doi{10.1117/12.506436}.

\bibitem[Depraetere et~al.(2007)Depraetere, Dahl, and Baldacchino]{depraetere2007world}
C~Depraetere, AL~Dahl, and G~Baldacchino.
\newblock A world of islands. an island studies reader.
\newblock 2007.

\bibitem[Dou et~al.(2023)Dou, Hong, Ciais, et~al.]{dou2023co2emissions}
X.~Dou, J.~Hong, P.~Ciais, et~al.
\newblock Near-real-time global gridded daily co2 emissions 2021.
\newblock \emph{Scientific Data}, 10:\penalty0 69, 2023.
\newblock \doi{10.1038/s41597-023-01963-0}.

\bibitem[Dyke \& Dyke(2001)Dyke and Dyke]{dyke2001introduction}
Phil~PG Dyke and PP~Dyke.
\newblock \emph{An introduction to Laplace transforms and Fourier series}.
\newblock Springer, 2001.

\bibitem[Flores et~al.(2022)Flores, Collins, Grineski, Amodeo, Porter, Sampson, and Wing]{Flores2022}
Aaron~B. Flores, Timothy~W. Collins, Sara~E. Grineski, Mike Amodeo, Jeremy~R. Porter, Christopher~C. Sampson, and Oliver Wing.
\newblock Federally overlooked flood risk inequities in houston, texas: Novel insights based on dasymetric mapping and state-of-the-art flood modeling.
\newblock \emph{Annals of the American Association of Geographers}, 2022.
\newblock \doi{10.1080/24694452.2022.2085656}.

\bibitem[Garcia-Silva et~al.(2018)Garcia-Silva, Gomez-Perez, Palma, Krystek, Mantovani, Foglini, Grande, Leo, Salvi, Trasati, Romaniello, Albani, Silvagni, Leone, Marelli, Albani, Lazzarini, Napier, Glaves, Aldridge, Meertens, Boler, Loescher, Laney, Genazzio, Crawl, and Altintas]{garciasilva2018enablingfairresearchearth}
Andres Garcia-Silva, Jose~Manuel Gomez-Perez, Raul Palma, Marcin Krystek, Simone Mantovani, Federica Foglini, Valentina Grande, Francesco~De Leo, Stefano Salvi, Elisa Trasati, Vito Romaniello, Mirko Albani, Cristiano Silvagni, Rosemarie Leone, Fulvio Marelli, Sergio Albani, Michele Lazzarini, Hazel~J. Napier, Helen~M. Glaves, Timothy Aldridge, Charles Meertens, Fran Boler, Henry~W. Loescher, Christine Laney, Melissa~A Genazzio, Daniel Crawl, and Ilkay Altintas.
\newblock Enabling fair research in earth science through research objects, 2018.
\newblock URL \url{https://arxiv.org/abs/1809.10617}.

\bibitem[Geneva \& Foster(2024)Geneva and Foster]{geneva_foster_2024_earth2studio}
Nicholas Geneva and Dallas Foster.
\newblock Nvidia earth2studio, 2024.
\newblock URL \url{https://github.com/NVIDIA/earth2studio}.

\bibitem[González(2009)]{Gonz_lez_2009}
Álvaro González.
\newblock Measurement of areas on a sphere using fibonacci and latitude–longitude lattices.
\newblock \emph{Mathematical Geosciences}, 42\penalty0 (1):\penalty0 49–64, November 2009.
\newblock ISSN 1874-8953.
\newblock \doi{10.1007/s11004-009-9257-x}.
\newblock URL \url{http://dx.doi.org/10.1007/s11004-009-9257-x}.

\bibitem[Grana et~al.(2021)Grana, Mukerji, and Doyen]{bookSeismic}
Dario Grana, Tapan Mukerji, and Philippe Doyen.
\newblock \emph{Seismic Reservoir Modeling: Theory, Examples, and Algorithms}.
\newblock 04 2021.
\newblock ISBN 9781119086215.
\newblock \doi{10.1002/9781119086215}.

\bibitem[Hersbach et~al.(1999)Hersbach, Bell, Berrisford, Hirahara, Hor{\'a}nyi, Mu{\~n}oz-Sabater, Nicolas, Peubey, Radu, Schepers, et~al.]{hersbach1999era5}
H~Hersbach, B~Bell, P~Berrisford, S~Hirahara, A~Hor{\'a}nyi, J~Mu{\~n}oz-Sabater, J~Nicolas, C~Peubey, R~Radu, D~Schepers, et~al.
\newblock The era5 global reanalysis, qj roy.
\newblock \emph{Meteor. Soc}, 146\penalty0 (730):\penalty0 1999--985, 1999.

\bibitem[Hillier et~al.(2023)Hillier, Wellmann, de~Kemp, Brodaric, Schetselaar, and B{\'e}dard]{hillier2023geoinr}
Michael Hillier, Florian Wellmann, Eric~A de~Kemp, Boyan Brodaric, Ernst Schetselaar, and Karine B{\'e}dard.
\newblock Geoinr 1.0: an implicit neural network approach to three-dimensional geological modelling.
\newblock \emph{Geoscientific Model Development}, 16\penalty0 (23):\penalty0 6987--7012, 2023.

\bibitem[Huffman et~al.(2014)Huffman, Bolvin, Braithwaite, Hsu, Joyce, and Xie]{huffman2014imerg}
G.~Huffman, D.~Bolvin, D.~Braithwaite, K.~Hsu, R.~Joyce, and P.~Xie.
\newblock Integrated multi-satellite retrievals for gpm (imerg), version 4.4.
\newblock Technical report, NASA's Precipitation Processing Center, 2014.
\newblock URL \url{ftp://arthurhou.pps.eosdis.nasa.gov/gpmdata/}.
\newblock Accessed 29 August, 2024.

\bibitem[Karger et~al.(2017)Karger, Conrad, Böhner, Kawohl, Kreft, Soria-Auza, Zimmermann, Linder, and Kessler]{karger2017climatologies}
D.~N. Karger, O.~Conrad, J.~Böhner, T.~Kawohl, H.~Kreft, R.~W. Soria-Auza, N.~E. Zimmermann, P.~Linder, and M.~Kessler.
\newblock Climatologies at high resolution for the earth land surface areas.
\newblock \emph{Scientific Data}, 4:\penalty0 170122, 2017.
\newblock \doi{10.1038/sdata.2017.122}.

\bibitem[Kemp et~al.(2022)Kemp, Xu, Depledge, Ebi, Gibbins, Kohler, Rockstr{\"o}m, Scheffer, Schellnhuber, Steffen, and Lenton]{kemp2022climate}
Luke Kemp, Chi Xu, Joanna Depledge, Kristie~L Ebi, Goodwin Gibbins, Timothy~A Kohler, Johan Rockstr{\"o}m, Marten Scheffer, Hans~Joachim Schellnhuber, Will Steffen, and Timothy~M Lenton.
\newblock Climate endgame: Exploring catastrophic climate change scenarios.
\newblock \emph{Proceedings of the National Academy of Sciences}, 119\penalty0 (34):\penalty0 e2108146119, 2022.
\newblock \doi{10.1073/pnas.2108146119}.
\newblock PMID: 35914185; PMCID: PMC9407216.

\bibitem[Kirichenko et~al.(2023)Kirichenko, Ibrahim, Balestriero, Bouchacourt, Vedantam, Firooz, and Wilson]{kirichenko2023understandingdetrimentalclassleveleffects}
Polina Kirichenko, Mark Ibrahim, Randall Balestriero, Diane Bouchacourt, Ramakrishna Vedantam, Hamed Firooz, and Andrew~Gordon Wilson.
\newblock Understanding the detrimental class-level effects of data augmentation, 2023.
\newblock URL \url{https://arxiv.org/abs/2401.01764}.

\bibitem[Liu et~al.(2024)Liu, Chen, Chen, Liu, An, Xia, and Wang]{liu2024efficientimplicitneuralrepresentation}
Xiang Liu, Jiahong Chen, Bin Chen, Zimo Liu, Baoyi An, Shu-Tao Xia, and Zhi Wang.
\newblock An efficient implicit neural representation image codec based on mixed autoregressive model for low-complexity decoding, 2024.
\newblock URL \url{https://arxiv.org/abs/2401.12587}.

\bibitem[Mai et~al.(2020)Mai, Janowicz, Yan, Zhu, Cai, and Lao]{mai2020multiscalerepresentationlearningspatial}
Gengchen Mai, Krzysztof Janowicz, Bo~Yan, Rui Zhu, Ling Cai, and Ni~Lao.
\newblock Multi-scale representation learning for spatial feature distributions using grid cells, 2020.
\newblock URL \url{https://arxiv.org/abs/2003.00824}.

\bibitem[Mai et~al.(2022)Mai, Janowicz, Hu, Gao, Yan, Zhu, Cai, and Lao]{Mai_2022}
Gengchen Mai, Krzysztof Janowicz, Yingjie Hu, Song Gao, Bo~Yan, Rui Zhu, Ling Cai, and Ni~Lao.
\newblock A review of location encoding for geoai: methods and applications.
\newblock \emph{International Journal of Geographical Information Science}, 36\penalty0 (4):\penalty0 639–673, January 2022.
\newblock ISSN 1362-3087.
\newblock \doi{10.1080/13658816.2021.2004602}.
\newblock URL \url{http://dx.doi.org/10.1080/13658816.2021.2004602}.

\bibitem[Mai et~al.(2023)Mai, Xuan, Zuo, He, Song, Ermon, Janowicz, and Lao]{mai2023sphere2vecgeneralpurposelocationrepresentation}
Gengchen Mai, Yao Xuan, Wenyun Zuo, Yutong He, Jiaming Song, Stefano Ermon, Krzysztof Janowicz, and Ni~Lao.
\newblock Sphere2vec: A general-purpose location representation learning over a spherical surface for large-scale geospatial predictions, 2023.
\newblock URL \url{https://arxiv.org/abs/2306.17624}.

\bibitem[Mallat(1999)]{mallat1999wavelet}
Stephane Mallat.
\newblock \emph{A wavelet tour of signal processing}.
\newblock Academic Press, 1999.

\bibitem[McEwen et~al.(2007)McEwen, Hobson, Mortlock, and Lasenby]{McEwen_2007}
Jason~D. McEwen, Michael~P. Hobson, Daniel~J. Mortlock, and Anthony~N. Lasenby.
\newblock Fast directional continuous spherical wavelet transform algorithms.
\newblock \emph{IEEE Transactions on Signal Processing}, 55\penalty0 (2):\penalty0 520–529, February 2007.
\newblock ISSN 1053-587X.
\newblock \doi{10.1109/tsp.2006.887148}.
\newblock URL \url{http://dx.doi.org/10.1109/TSP.2006.887148}.

\bibitem[Mildenhall et~al.(2020)Mildenhall, Srinivasan, Tancik, Barron, Ramamoorthi, and Ng]{mildenhall2020nerfrepresentingscenesneural}
Ben Mildenhall, Pratul~P. Srinivasan, Matthew Tancik, Jonathan~T. Barron, Ravi Ramamoorthi, and Ren Ng.
\newblock Nerf: Representing scenes as neural radiance fields for view synthesis, 2020.
\newblock URL \url{https://arxiv.org/abs/2003.08934}.

\bibitem[Molaei et~al.(2023)Molaei, Aminimehr, Tavakoli, Kazerouni, Azad, Azad, and Merhof]{molaei2023implicitneuralrepresentationmedical}
Amirali Molaei, Amirhossein Aminimehr, Armin Tavakoli, Amirhossein Kazerouni, Bobby Azad, Reza Azad, and Dorit Merhof.
\newblock Implicit neural representation in medical imaging: A comparative survey, 2023.
\newblock URL \url{https://arxiv.org/abs/2307.16142}.

\bibitem[Munday \& Washington(2018)Munday and Washington]{Munday2018}
C.~Munday and R.~Washington.
\newblock Systematic climate model rainfall biases over southern africa: Links to moisture circulation and topography.
\newblock \emph{Journal of Climate}, 31\penalty0 (18):\penalty0 7533--7548, 2018.
\newblock \doi{10.1175/JCLI-D-18-0008.1}.
\newblock URL \url{https://doi.org/10.1175/JCLI-D-18-0008.1}.

\bibitem[Murphy et~al.(2006)Murphy, Ardanuy, Deluccia, Clement, and Schueler]{Murphy2006}
R.~E. Murphy, Phillip Ardanuy, Frank~J. Deluccia, J.~E. Clement, and Carl~F. Schueler.
\newblock \emph{The Visible Infrared Imaging Radiometer Suite}, pp.\  199--223.
\newblock Springer Berlin Heidelberg, Berlin, Heidelberg, 2006.
\newblock ISBN 978-3-540-37293-6.
\newblock \doi{10.1007/978-3-540-37293-6_11}.
\newblock URL \url{https://doi.org/10.1007/978-3-540-37293-6_11}.

\bibitem[{NASA Shuttle Radar Topography Mission}(2013)]{SRTM2013}
{NASA Shuttle Radar Topography Mission}.
\newblock Shuttle radar topography mission (srtm) global.
\newblock \url{https://doi.org/10.5069/G9445JDF}, 2013.
\newblock URL \url{https://doi.org/10.5069/G9445JDF}.
\newblock Distributed by OpenTopography.

\bibitem[{OCO-2 Science Team/Michael Gunson, Annmarie Eldering}(2020)]{OCO2}
{OCO-2 Science Team/Michael Gunson, Annmarie Eldering}.
\newblock {OCO-2 Level 2 bias-corrected XCO2 and other select fields from the full-physics retrieval aggregated as daily files}, 2020.
\newblock URL \url{https://disc.gsfc.nasa.gov/datasets/OCO2_L2_Lite_FP_10r/summary}.

\bibitem[O'Dell et~al.(2018)O'Dell, Eldering, Wennberg, Crisp, Gunson, Fisher, Frankenberg, Kiel, Lindqvist, Mandrake, Merrelli, Natraj, Nelson, Osterman, Payne, Taylor, Wunch, Drouin, Oyafuso, Chang, McDuffie, Smyth, Baker, Basu, Chevallier, Crowell, Feng, Palmer, Dubey, García, Griffith, Hase, Iraci, Kivi, Morino, Notholt, Ohyama, Petri, Roehl, Sha, Strong, Sussmann, Te, Uchino, and Velazco]{ODell2018}
C.~W. O'Dell, A.~Eldering, P.~O. Wennberg, D.~Crisp, M.~R. Gunson, B.~Fisher, C.~Frankenberg, M.~Kiel, H.~Lindqvist, L.~Mandrake, A.~Merrelli, V.~Natraj, R.~R. Nelson, G.~B. Osterman, V.~H. Payne, T.~E. Taylor, D.~Wunch, B.~J. Drouin, F.~Oyafuso, A.~Chang, J.~McDuffie, M.~Smyth, D.~F. Baker, S.~Basu, F.~Chevallier, S.~M.~R. Crowell, L.~Feng, P.~I. Palmer, M.~Dubey, O.~E. García, D.~W.~T. Griffith, F.~Hase, L.~T. Iraci, R.~Kivi, I.~Morino, J.~Notholt, H.~Ohyama, C.~Petri, C.~M. Roehl, M.~K. Sha, K.~Strong, R.~Sussmann, Y.~Te, O.~Uchino, and V.~A. Velazco.
\newblock Improved retrievals of carbon dioxide from orbiting carbon observatory-2 with the version 8 acos algorithm.
\newblock \emph{Atmospheric Measurement Techniques}, 11\penalty0 (12):\penalty0 6539--6576, 2018.
\newblock \doi{10.5194/amt-11-6539-2018}.

\bibitem[{OpenStreetMap contributors}(2017)]{OpenStreetMap}
{OpenStreetMap contributors}.
\newblock {Planet dump retrieved from https://planet.osm.org }.
\newblock \url{ https://www.openstreetmap.org }, 2017.

\bibitem[Palecki et~al.(2013)Palecki, Lawrimore, Leeper, Bell, Embler, and Casey]{palecki2013uscrn}
M.A. Palecki, J.H. Lawrimore, R.D. Leeper, J.E. Bell, S.~Embler, and N.~Casey.
\newblock U.s. climate reference network products, 2013.
\newblock Accessed 29 August, 2024.

\bibitem[Raissi et~al.(2017)Raissi, Perdikaris, and Karniadakis]{raissi2017physicsinformeddeeplearning}
Maziar Raissi, Paris Perdikaris, and George~Em Karniadakis.
\newblock Physics informed deep learning (part i): Data-driven solutions of nonlinear partial differential equations, 2017.
\newblock URL \url{https://arxiv.org/abs/1711.10561}.

\bibitem[Rußwurm et~al.(2024)Rußwurm, Klemmer, Rolf, Zbinden, and Tuia]{rußwurm2024geographiclocationencodingspherical}
Marc Rußwurm, Konstantin Klemmer, Esther Rolf, Robin Zbinden, and Devis Tuia.
\newblock Geographic location encoding with spherical harmonics and sinusoidal representation networks, 2024.
\newblock URL \url{https://arxiv.org/abs/2310.06743}.

\bibitem[Sanz et~al.(2006)Sanz, Herranz, Lopez-Caniego, and Argueso]{sanz2006waveletssphereapplicationdetection}
J.~L. Sanz, D.~Herranz, M.~Lopez-Caniego, and F.~Argueso.
\newblock Wavelets on the sphere. application to the detection problem, 2006.
\newblock URL \url{https://arxiv.org/abs/astro-ph/0609351}.

\bibitem[Saragadam et~al.(2023)Saragadam, LeJeune, Tan, Balakrishnan, Veeraraghavan, and Baraniuk]{saragadam2023wirewaveletimplicitneural}
Vishwanath Saragadam, Daniel LeJeune, Jasper Tan, Guha Balakrishnan, Ashok Veeraraghavan, and Richard~G. Baraniuk.
\newblock Wire: Wavelet implicit neural representations, 2023.
\newblock URL \url{https://arxiv.org/abs/2301.05187}.

\bibitem[Sitzmann et~al.(2020)Sitzmann, Martel, Bergman, Lindell, and Wetzstein]{sitzmann2020implicit}
Vincent Sitzmann, Julien Martel, Alexander Bergman, David Lindell, and Gordon Wetzstein.
\newblock Implicit neural representations with periodic activation functions.
\newblock \emph{Advances in neural information processing systems}, 33:\penalty0 7462--7473, 2020.

\bibitem[Sorooshian et~al.(2014)Sorooshian, Hsu, Braithwaite, Ashouri, and {NOAA CDR Program}]{sorooshian2014persiann}
Soroosh Sorooshian, Kuolin Hsu, Dan Braithwaite, Hamed Ashouri, and {NOAA CDR Program}.
\newblock Noaa climate data record (cdr) of precipitation estimation from remotely sensed information using artificial neural networks (persiann-cdr), version 1 revision 1, 2014.
\newblock Accessed 29 August, 2024.

\bibitem[Stone et~al.(2024)Stone, Ravikumar, Bulpitt, and Hogg]{stone2024epistemicuncertaintyweightedlossvisual}
Rebecca~S Stone, Nishant Ravikumar, Andrew~J Bulpitt, and David~C Hogg.
\newblock Epistemic uncertainty-weighted loss for visual bias mitigation, 2024.
\newblock URL \url{https://arxiv.org/abs/2204.09389}.

\bibitem[Taylor et~al.(2023)Taylor, O'Dell, Eldering, Crisp, Fisher, Gunson, Basilio, Kronk, Kiel, Kuai, Osterman, Nelson, Payne, Wennberg, Wunch, and Ziegler]{Taylor2023}
T.~E. Taylor, C.~W. O'Dell, A.~Eldering, D.~Crisp, B.~M. Fisher, M.~R. Gunson, R.~R. Basilio, F.~Kronk, M.~Kiel, L.~Kuai, G.~B. Osterman, R.~R. Nelson, V.~H. Payne, P.~O. Wennberg, D.~Wunch, and J.~Ziegler.
\newblock The oco-3 retrievals and data products.
\newblock \emph{Atmospheric Measurement Techniques}, 16\penalty0 (16):\penalty0 3889--3921, 2023.
\newblock \doi{10.5194/amt-16-3889-2023}.

\bibitem[Tseng et~al.(2022)Tseng, Kerner, and Rolnick]{tseng2022timltaskinformedmetalearningagriculture}
Gabriel Tseng, Hannah Kerner, and David Rolnick.
\newblock Timl: Task-informed meta-learning for agriculture, 2022.
\newblock URL \url{https://arxiv.org/abs/2202.02124}.

\bibitem[Xu et~al.(2022)Xu, Wang, Jiang, Fan, and Wang]{xu2022signalprocessingimplicitneural}
Dejia Xu, Peihao Wang, Yifan Jiang, Zhiwen Fan, and Zhangyang Wang.
\newblock Signal processing for implicit neural representations, 2022.
\newblock URL \url{https://arxiv.org/abs/2210.08772}.

\end{thebibliography}
\bibliographystyle{iclr2025_conference}

\appendix
\section{Appendix}

\subsection{\textsc{FAIR-Earth} Details}
\label{APP:FAIR-Earth Details}

\begin{longtable}{|p{0.12\textwidth}|p{0.4\textwidth}|p{0.17\textwidth}|p{0.15\textwidth}|}
\caption{FAIR-Earth Components}\label{tab:dataset-components} \\
\hline
\textbf{Category} & \textbf{Description} & \textbf{Misc.} & \textbf{Source} \\
\hline
\endfirsthead
\multicolumn{4}{l}{\textit{Continued from previous page}} \\
\hline
\textbf{Label} & \textbf{Description} & \textbf{Misc.} & \textbf{Source} \\
\hline
\endhead
\hline
\multicolumn{4}{r}{\textit{Continued on next page}} \\
\endfoot
\hline
\endlastfoot
Land-Sea & Binary and continuous data, based on bathymetric data on percent water surface coverage for every grid point. & Metadata includes: islands, coastlines, landmass sizes, coast distance. & \citep{huffman2014imerg} \\
\hline
Population & Population density data, derived from combination of spatial distribution with national census and population register data. & Errors in Egypt and Greenland are flagged, and tentatively smoothed via nearest-neighbor interpolation. & \citep{ciesin2018gpw} \\
\hline
Precipitation & Global accumulated precipitation, based on composite of measurements and modeling. & Monthly Resolution & \citep{beck2019mswep} \\
\hline
Temperature & Air surface temperature, based on satellite measurements and reflectance modeling. & Monthly Resolution & \citep{karger2017climatologies} \\
\hline
Emissions & Measured and modeled CO2 emissions from the OCO-2 satellite. & Daily Resolution & \citep{dou2023co2emissions} \\
\hline
\end{longtable}

\paragraph{Metadata Details}
\label{APP:MetadataDetails}
In particular, for subgroups with ill-defined thresholds (e.g., classification of islands and coastlines), we provide flexibility with adjustable thresholds. This feature allows researchers to fine-tune their analyses based on specific definitions or needs, which can vary depending on the research question or application domain. For the purposes of our analyses, islands are defined as landmasses with size under $30,000$ sq. miles, encapsulating most of the ``minor islands'' as defined by \citet{depraetere2007world}. 

Second, we incorporate {\em demographics and geographical boundaries} leveraging the Gridded Population of the World, Version 4 (GPWv4) population dataset \citep{ciesin_gpwv4_2018}, which integrates censuses, population registers, and spatial distributions, we synthesize population density data for each point in the grid. Then, leveraging high-resolution NaturalEarth \citep{NaturalEarth} data, we label each grid point with appropriate country labels. These labels are essential for equal representation of both densely and sparsely populated areas, preventing the overrepresentation of urban centers or the underrepresentation of rural regions, a common bias seen in survey and modeling tasks.

\subsubsection{Dataset Figures}
\setlength{\belowcaptionskip}{-10pt}
\begin{figure}[t!]
\centering
\includegraphics[width=1.0\textwidth]{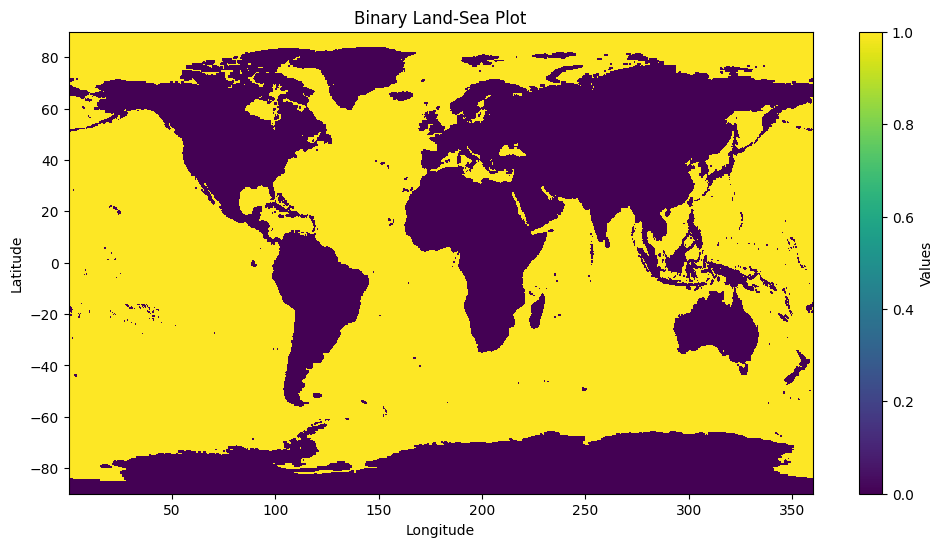}
\caption{Binary land-sea}
\label{fig:image1}
\end{figure}

\begin{figure}[t!]
\centering
\includegraphics[width=1.0\textwidth]{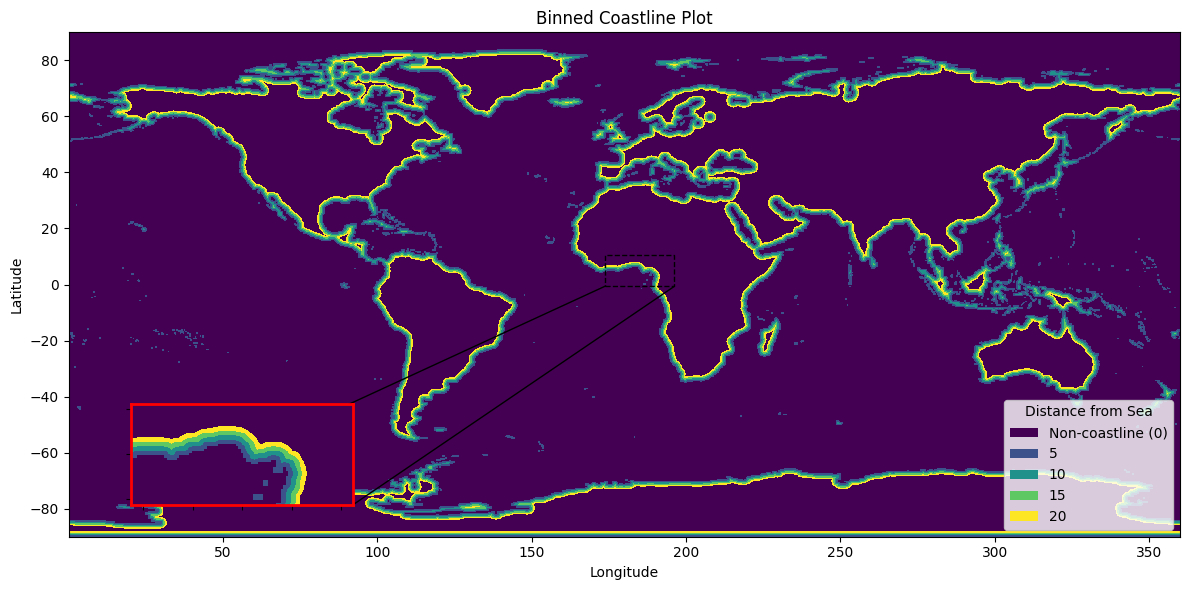}
\caption{Coastline}
\label{fig:image3}
\end{figure}

\begin{figure}[t!]
\centering
\includegraphics[width=1.0\textwidth]{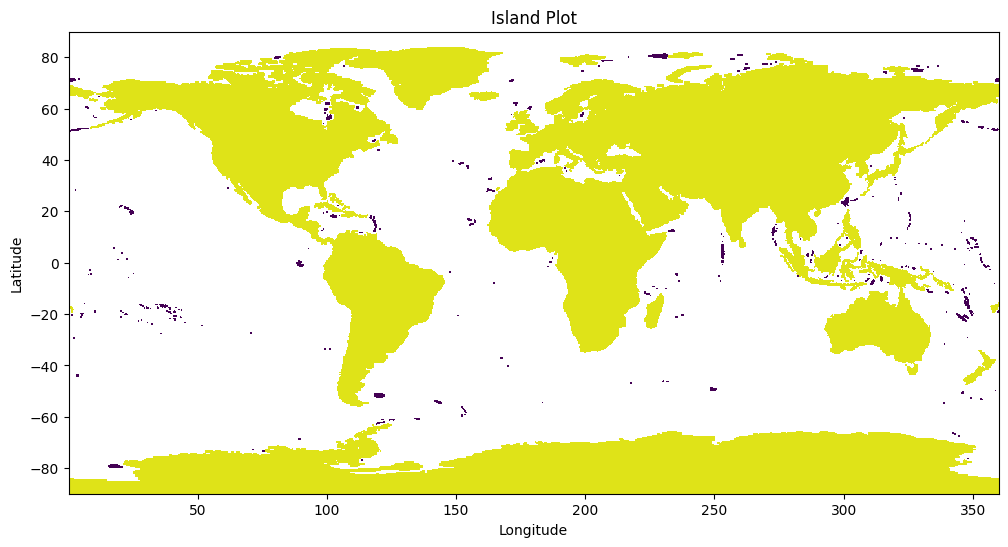}
\caption{Islands (in dark purple)}
\label{fig:image4}
\end{figure}

\begin{figure}[t!]
\centering
\includegraphics[width=1.0\textwidth]{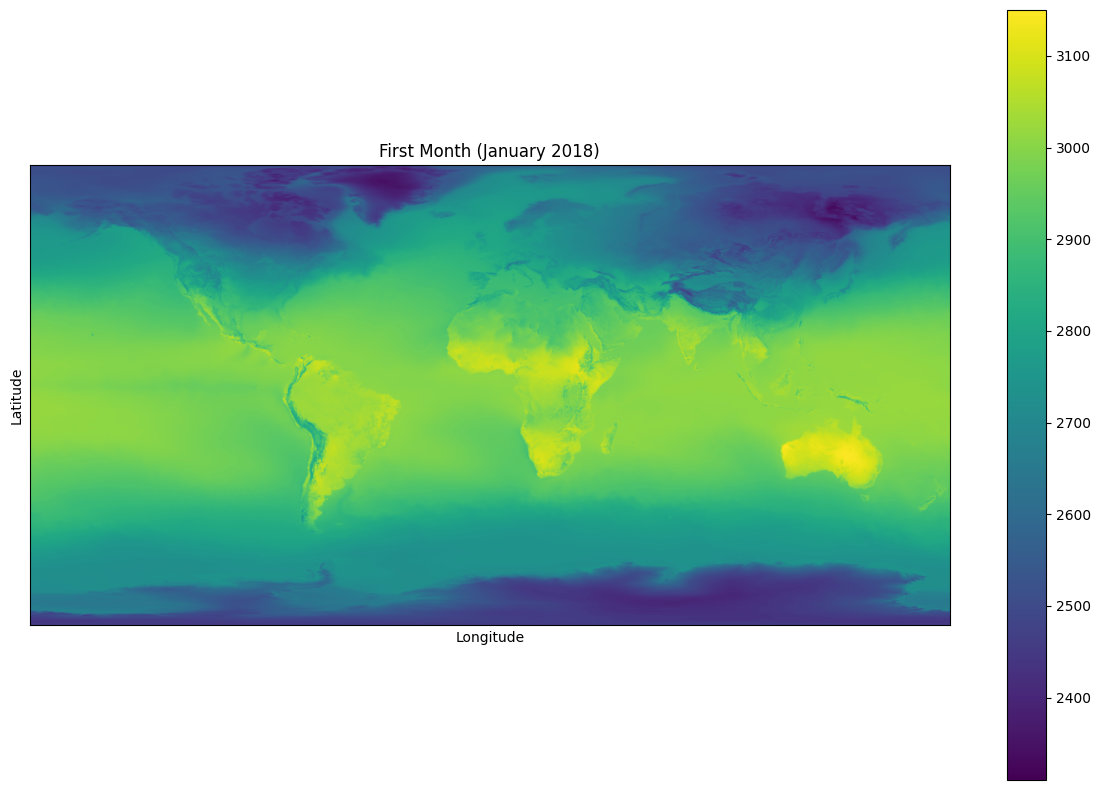}
\caption{Air surface temperature plot (Jan. 2018)}
\label{fig:image5}
\end{figure}

\begin{figure}[t!]
\centering
\includegraphics[width=1.0\textwidth]{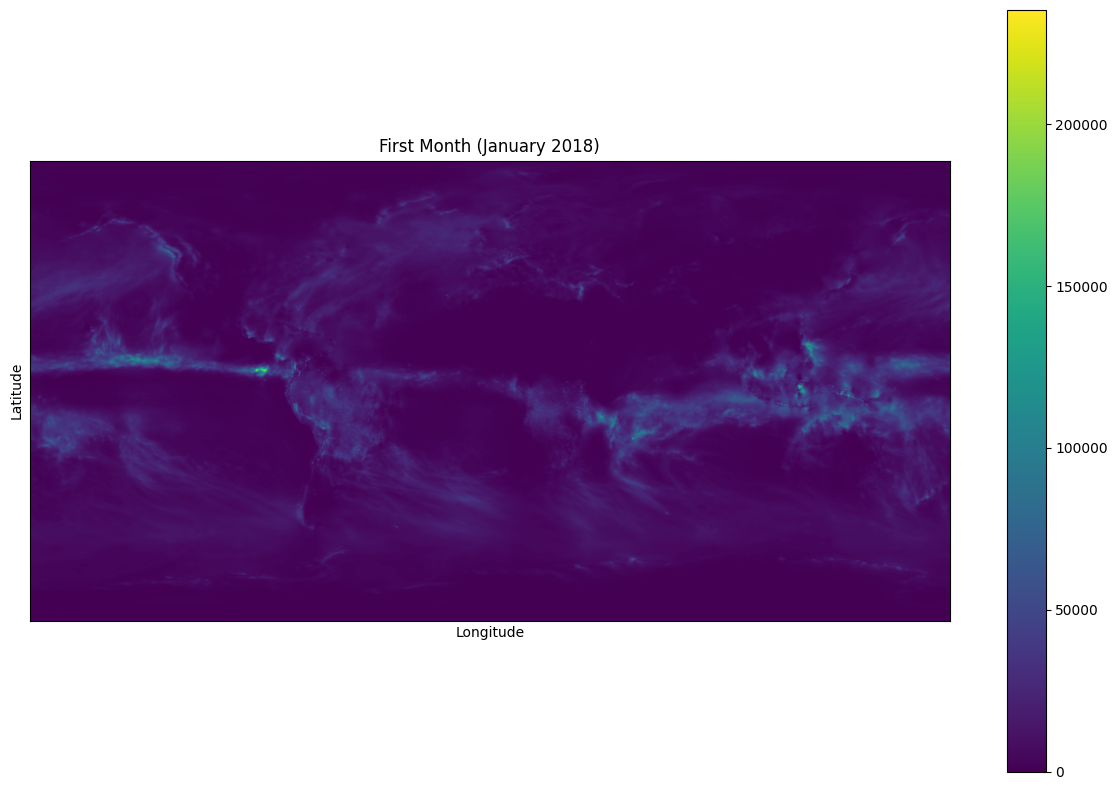}
\caption{Cumulative precipitation plot (Jan. 2018)}
\label{fig:image6}
\end{figure}

\begin{figure}[t!]
\centering
\includegraphics[width=1.0\textwidth]{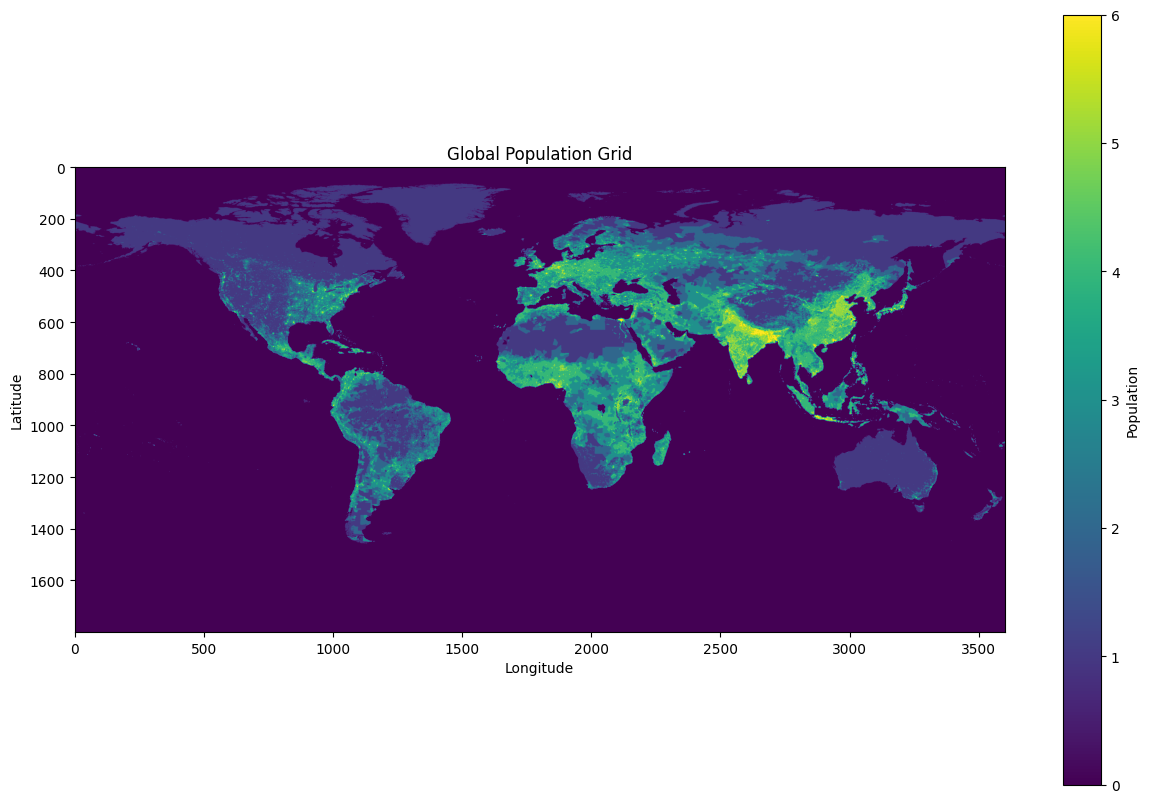}
\caption{Global population plot}
\label{fig:image7}
\end{figure}

\begin{figure}[t!]
\centering
\includegraphics[width=1.0\textwidth]{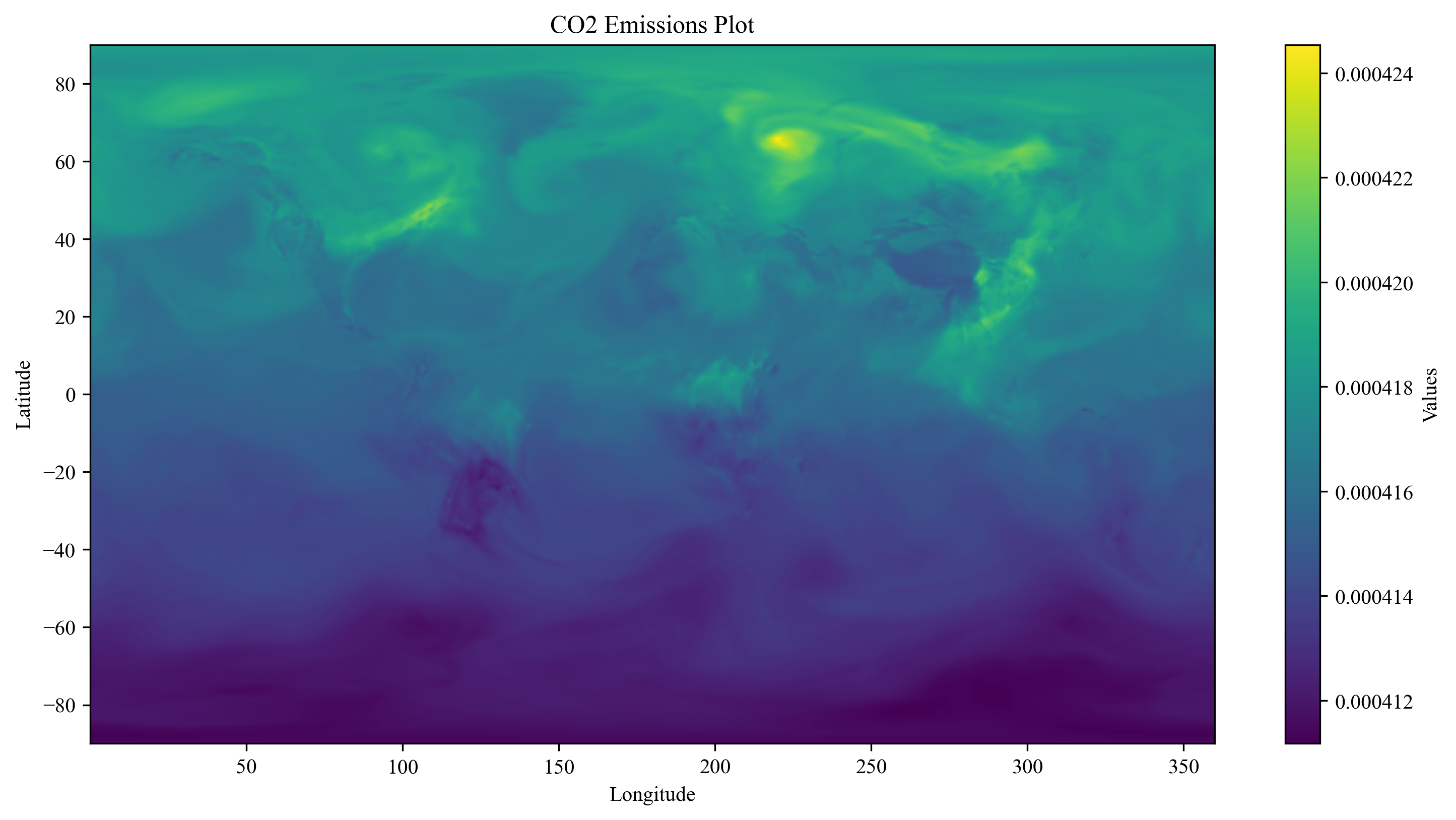}
\caption{CO2 emissions plot}
\label{fig:image8}
\end{figure}

\begin{figure}[t!]
\centering
\includegraphics[width=1.0\textwidth]{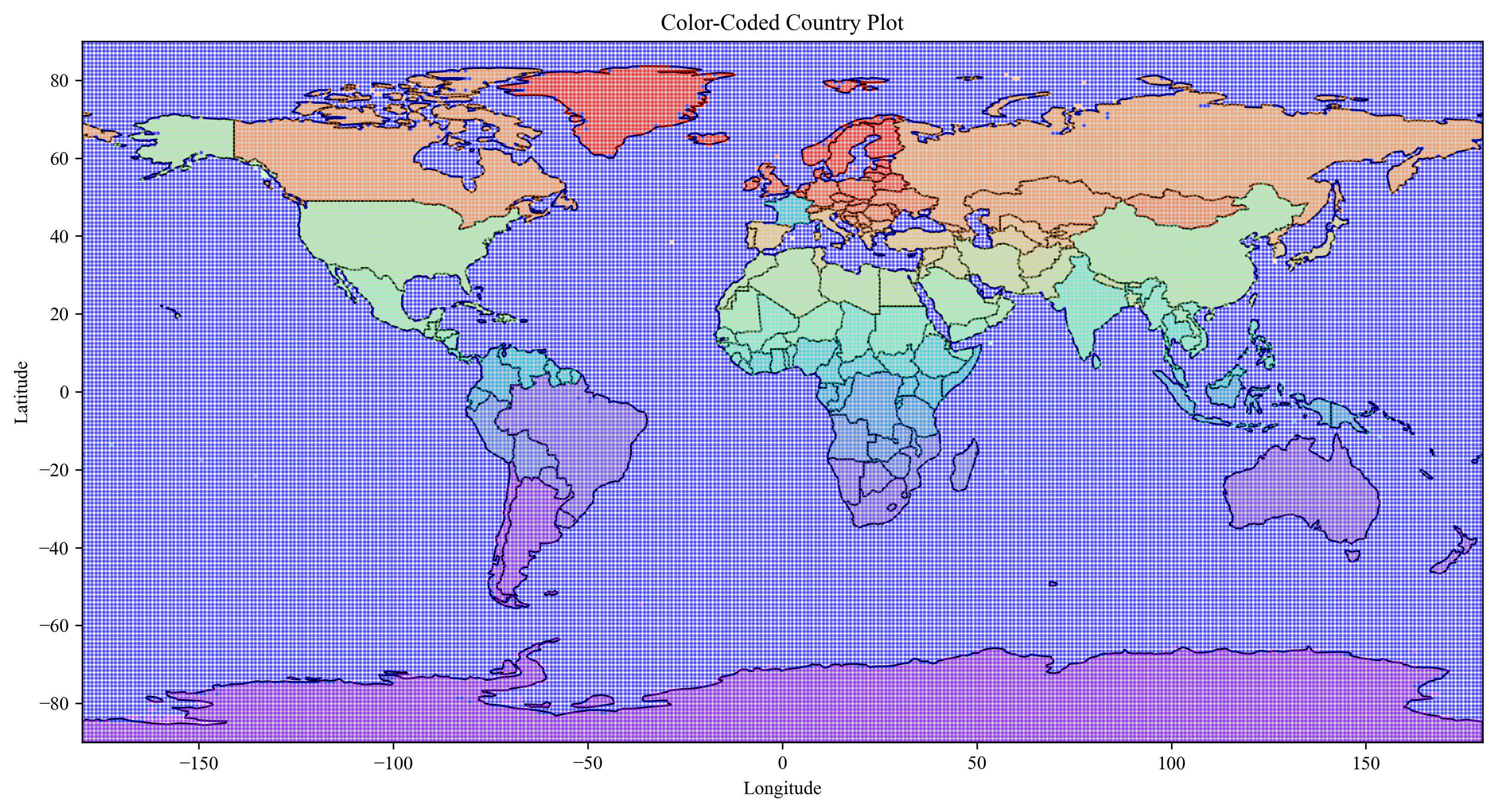}
\caption{Color-coded country plot}
\label{fig:image9}
\end{figure}

\clearpage
\newpage

\subsection{Training Specifications}
\label{APP:Training Specifications}
\begin{figure}[t!]
    \centering
    \begin{tabular}{cc}
        \includegraphics[width=0.48\textwidth]{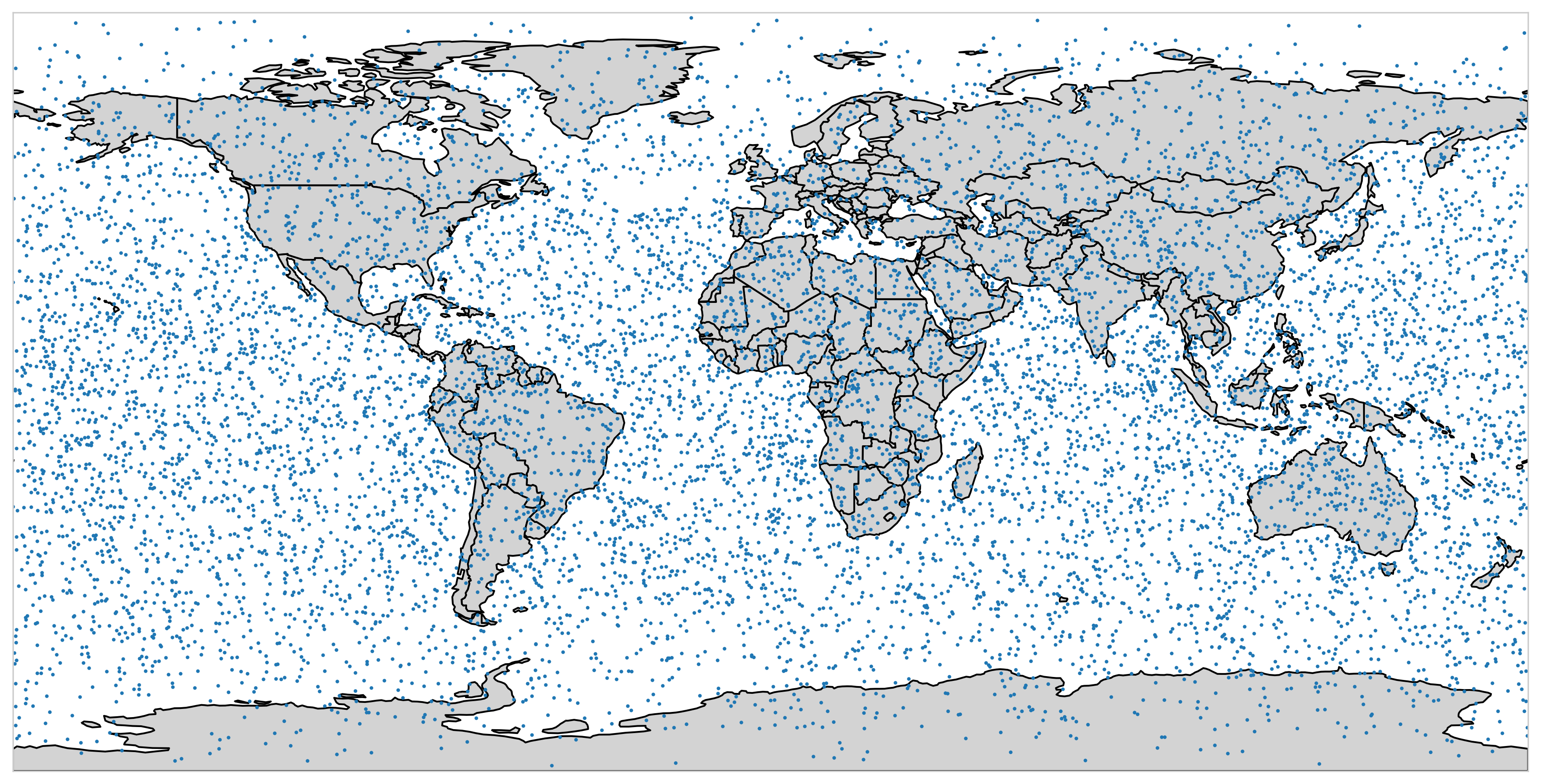} &
        \includegraphics[width=0.48\textwidth]{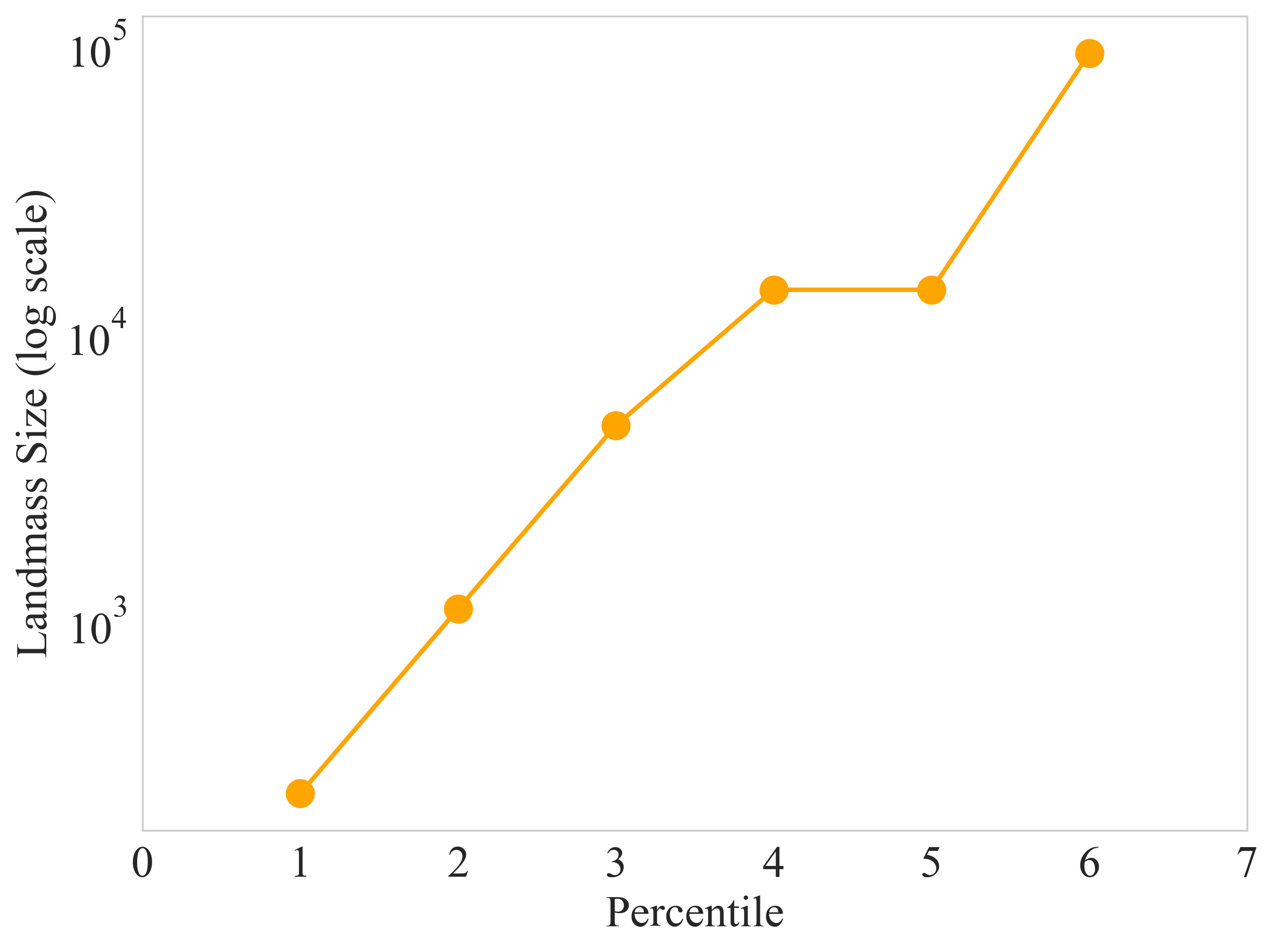} \\
        a) Sampled Points ($N=10000$) &
        b) Landmass Size Percentiles for $N_{\text{land}}=3233$ \\
    \end{tabular}
    \caption{Coverage of different landmass sizes using uniform sampling. At $N=10000$, at least 64 points are sampled on landmasses of size less than $40,000$ square miles, roughly the size of the main island of Cuba.}
    \label{fig:pointDistribution}
\end{figure}

\paragraph{Land-Sea Binary Classification Training} We conducted a comprehensive ablation study to examine model trends on the land-sea binary classification task. The study was performed using a grid-search, and relevant parameters are available in \cref{tab:grid-search-params}. Note that the parameter sizes for encodings were chosen to be similar, to control for any effects from simple increase in embedding space.

\begin{table}[t!]
\centering
\caption{Grid search and encoding parameters}
\label{tab:grid-search-params}
\small
\begin{tabular}{p{3cm}p{4.5cm}p{4cm}}
\hline
\textbf{Parameter} & \textbf{Values} & \textbf{Applicable Encoding} \\
\hline
Training Samples & \{5000, 10000, 15000, 20000\} & Both \\
Weight Decay & \{1e-5, 1e-4, 1e-3\} & All \\
Maximum Scale & \{3, 4, 5\} & \textsc{SW} \\
Maximum Rotations & \{50, 90, 130, 170\} & \textsc{SW} \\
Legendre Polynomials & \{5, 7, 10, 12, 15, 17, 20, 22, 25\} & \textsc{SH} \\
K Value & 6 & \textsc{SW} \\
Scale Factor & 1 & \textsc{SW} \\
Minimum Radius & \{45, 90\} & \textsc{Theory}, \\\mbox{} & & \textsc{Grid and Sphere} \\
Frequency Numbers & \{16, 32, 64\} & \textsc{Theory}, \\\mbox{} & & \textsc{Grid and Sphere} \\
Batch Size & 2048 & All \\
Learning Rate & 1e-4 & All \\
Maximum Epochs & 500 & All \\
\hline
\end{tabular}
\end{table}

To initialize training, we set identical data generation seeds for each model. Validation data was sampled similar to \cref{fig:pointDistribution} , but with $N_{\text{validation}}=0.2 * N_{\text{training}}$ and different initialization seeds.

\begin{figure}[t!]
   \centering
   \includegraphics[width=1.0\textwidth]{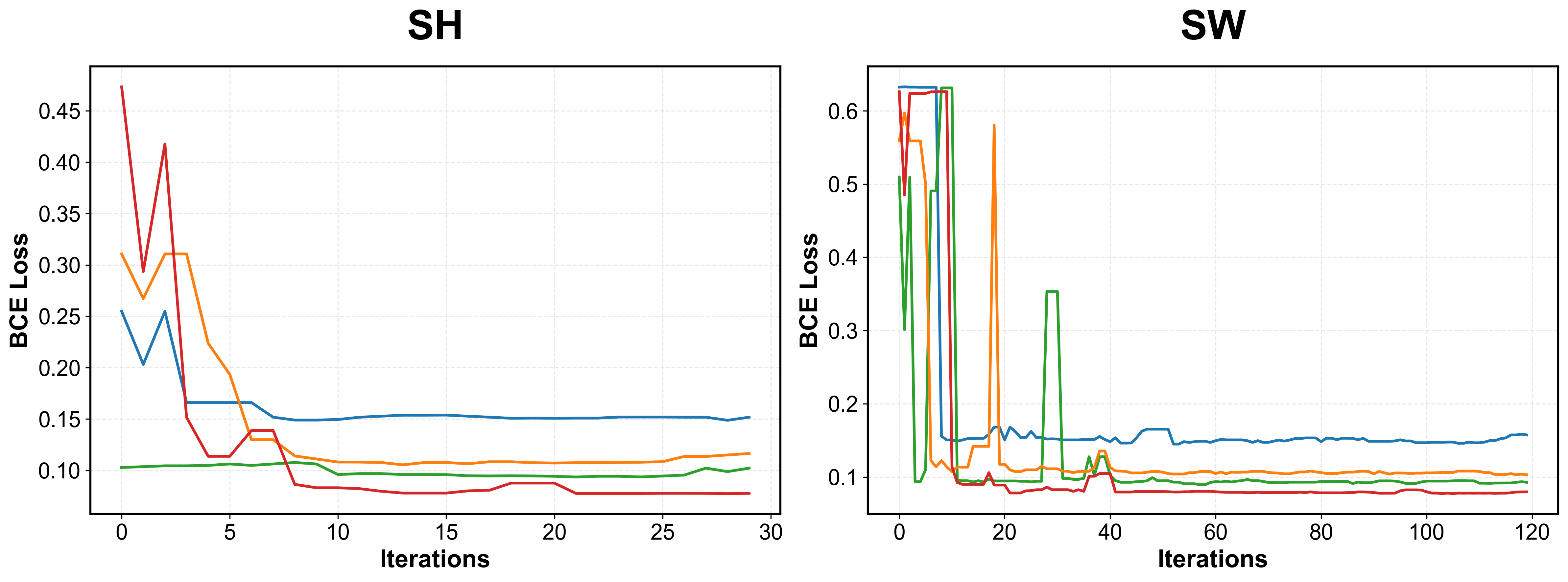}
   \caption{Fine tuning dynamics for different training resolutions, showing 5-iteration median window. Despite the larger parameter space, fine-tuning of \textsc{Spherical Wavelet} converges at a comparable, slightly slower rate than \textsc{Spherical Harmonic.}}
   \label{fig:fine-tuning-dynamics}
\end{figure}

For fine-tuning, we perform a similar grid search over a similar continuous range of variables. The specific parameters are shown in \cref{tab:hyperparameter-ranges}. Due to the extra parameterization of \textsc{Spherical Wavelet}, we quadruple the number of trials to 120, compared to 30 for all other encodings (\cref{fig:fine-tuning-dynamics}). 

\begin{table}[t!]
\centering
\caption{Hyperparameter ranges for model optimization}
\label{tab:hyperparameter-ranges}
\begin{tabular}{llll}
\textbf{Category} & \textbf{Hyperparameter} & \textbf{Range} & \textbf{Step} \\
\hline
\multirow{2}{*}{Architecture} & Hidden Dimension & 32 -- 96 & 32 \\
 & Number of Layers & 1 -- 3 & 1 \\
\hline
\multirow{4}{*}{\textsc{Spherical Wavelet}} & Maximum Scale & 2 -- 5 & 1 \\
 & Maximum Rotations & 20 -- 200 & 40 \\
 & K Value & 4 -- 10 & - \\
 & Scale Factor & 0.75 -- 1.25 & - \\
\hline
\textsc{Spherical Harmonic} & Legendre Polynomial Degree & 10 -- 30 & - \\
\hline
\multirow{2}{*}{Training} & Learning Rate & $10^{-4}$ -- $10^{-1}$ & log scale \\
 & Weight Decay & $10^{-8}$ -- $10^{-1}$ & log scale \\
\hline
\end{tabular}
\end{table}

\paragraph{Precipitation and Temperature Training} Training over these environmental signals was nearly identical to training on the land-sea binary classification task. We note a couple of minor discrepancies:
\begin{itemize}
    \item Mean-squared error loss was applied to a normalized version of the data, as opposed to binary cross-entropy.
    \item For the sake of precision and brevity, all precipitation losses in the paper are increased by a magnitude of 10. We emphasize that this doesn't affect our analyses.
\end{itemize}


\subsection{Miscellaneous Figures}
\label{APP:Misc Figures}

\begin{figure}[t!]
    \centering
    \begin{tabular}{ccc}
        \includegraphics[width=0.3\textwidth]{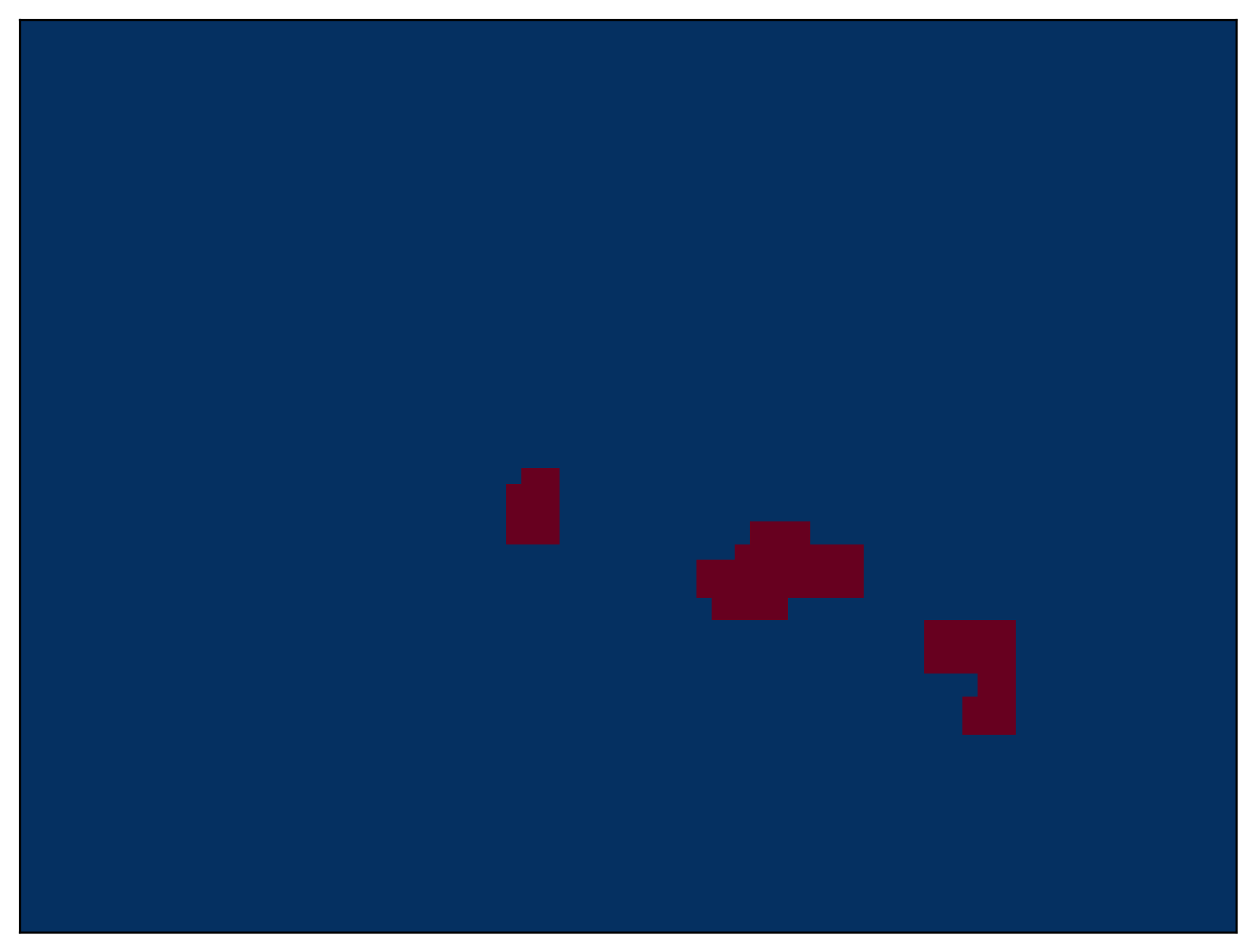} &
        \includegraphics[width=0.3\textwidth]{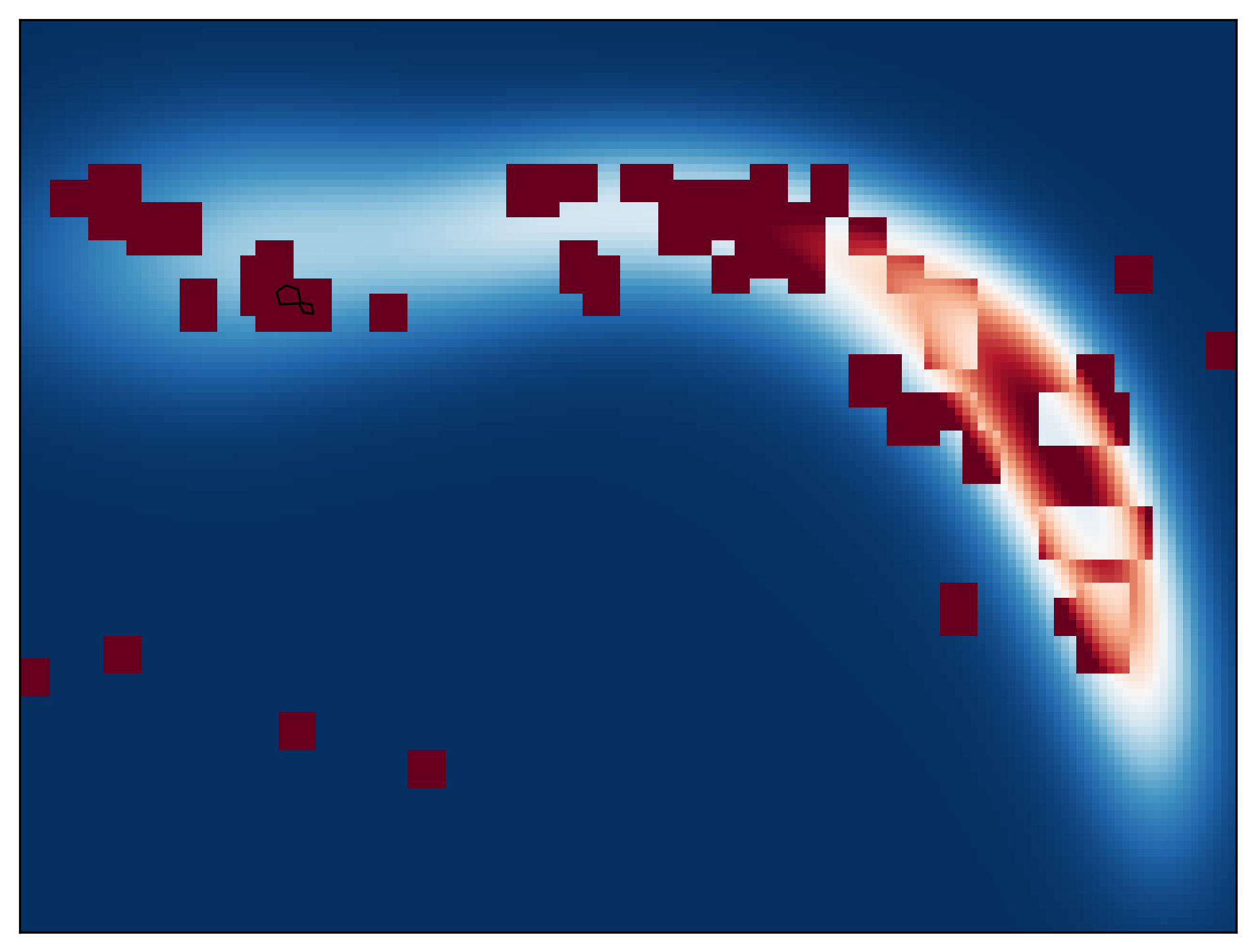} &
        \includegraphics[width=0.3\textwidth]{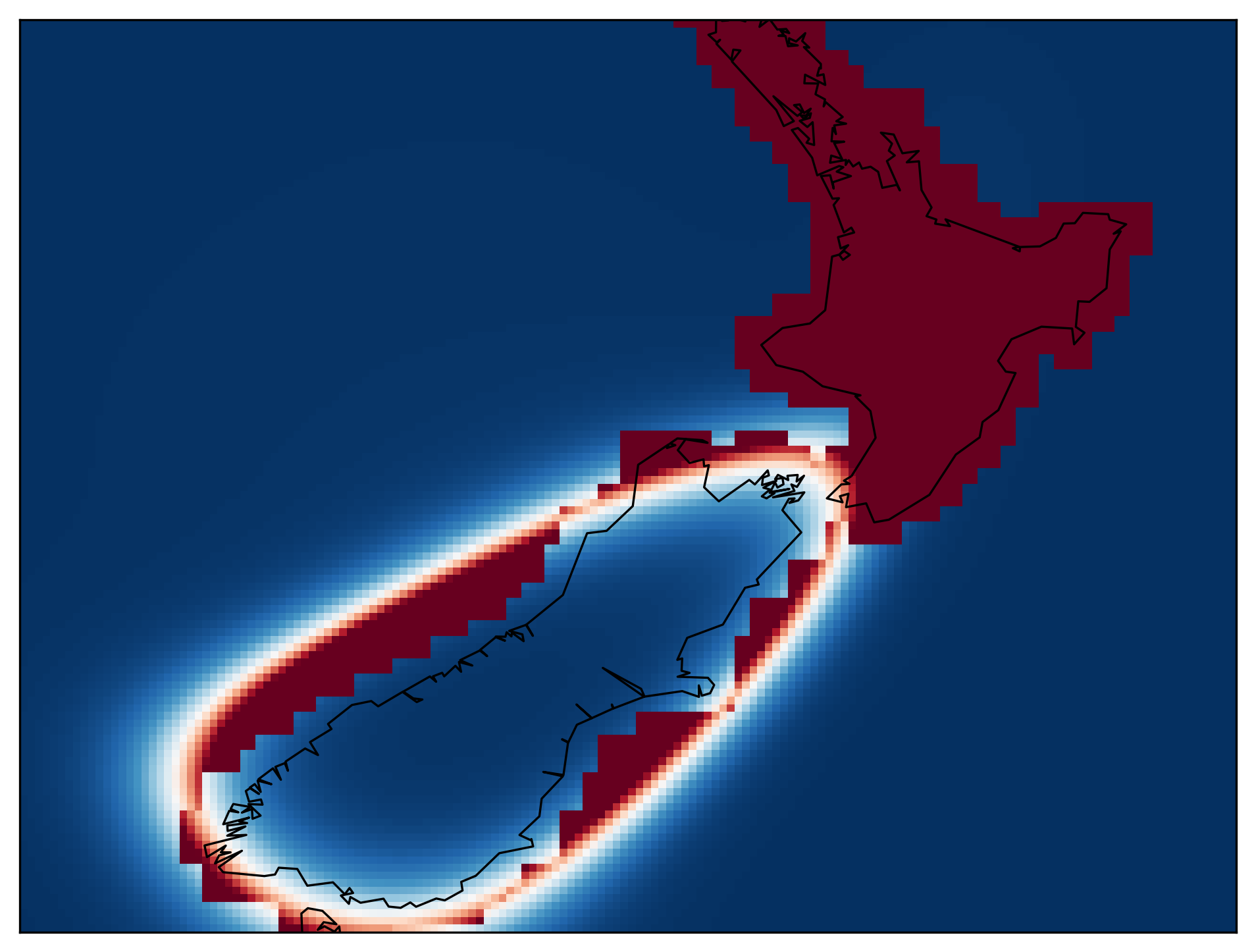} \\
    \end{tabular}
    \caption{Zoomed-in inset plots for main diagram}
    \label{fig:threeImages}
\end{figure}

\vspace{1em}

\begin{figure}[t!]
    \centering
    \includegraphics[width=0.25\textwidth]{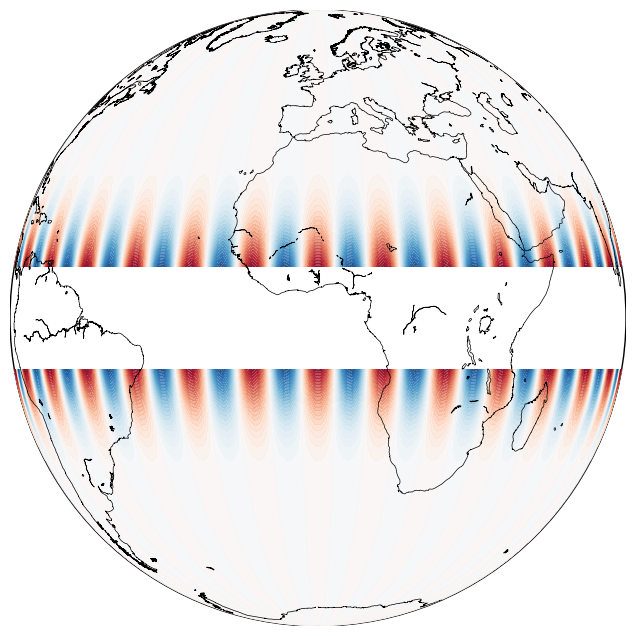}
    \caption{At higher frequencies ($L=30$), the closed-form calculations provided in \citet{rußwurm2024geographiclocationencodingspherical} run into numerical error.}
    \label{fig:legendreNans}
\end{figure}

\vspace{1em}

\begin{figure}[t!]
\centering
\includegraphics[width=0.8\textwidth]{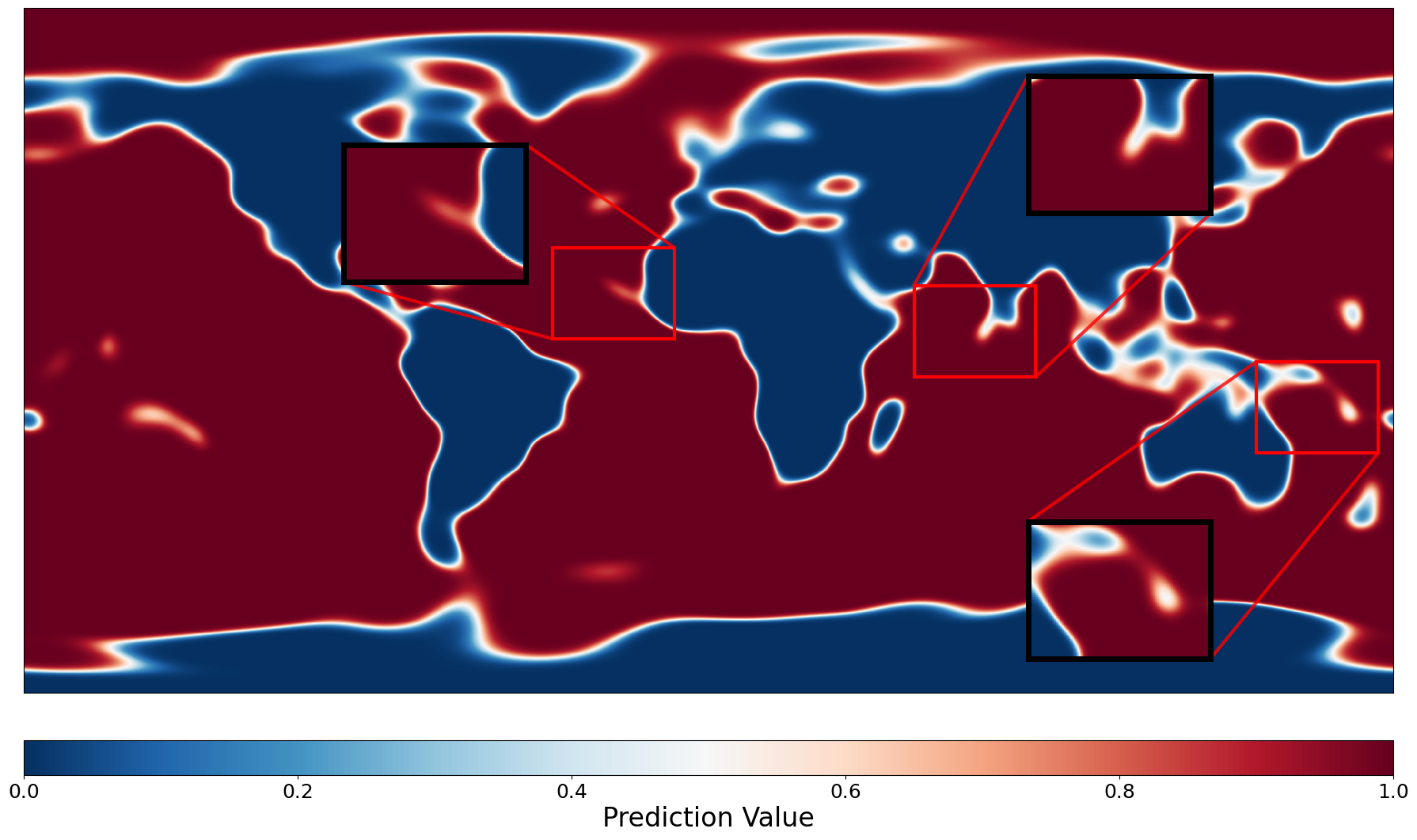}
\caption{\textsc{Spherical Harmonic} encoding prediction plot}
\label{fig:SHprediction}
\end{figure}

\begin{figure}[t!]
\centering
\includegraphics[width=0.8\textwidth]{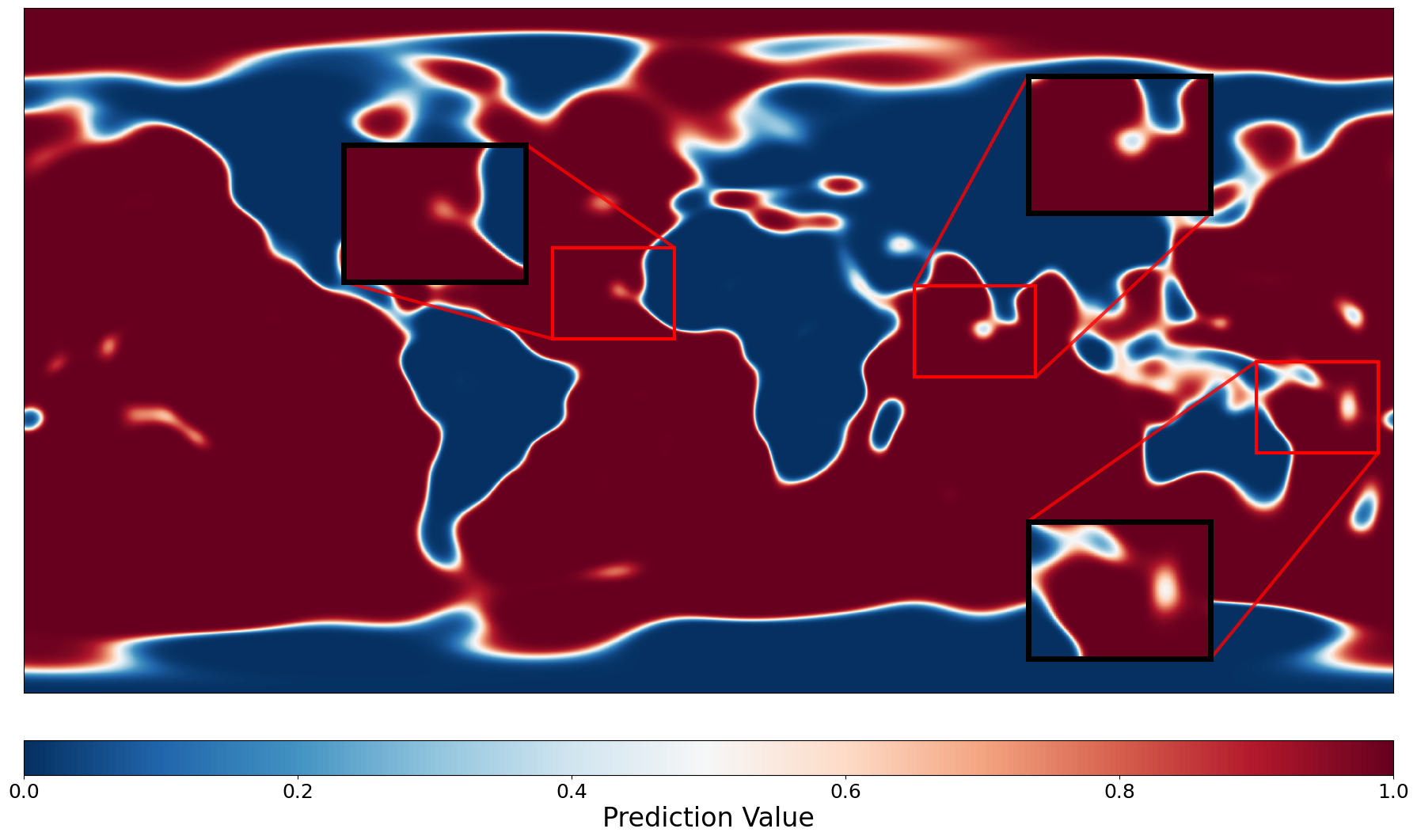}
\caption{\textsc{Spherical Wavelet} encoding prediction plot}
\label{fig:SWprediction}
\end{figure}

\clearpage

\begin{table}[t!]
\centering
\caption{Comparison of existing location encodings}
\begin{tabular}{l|c|c|c}
\hline
Encoding & Resolution-Adaptive & Basis Type & Localized Basis \\
\hline
\textsc{Spherical Wavelet} & \checkmark & Wavelet & \checkmark \\
\textsc{Spherical Harmonic} & \checkmark & Harmonic & $\times$  \\
\textsc{Cartesian3D} & $\times$ & Coordinate & $\times$ \\
\textsc{Theory} & \checkmark & Sinusoidal + Coordinate & $\times$ \\
\textsc{Grid and Sphere} & \checkmark & Sinusoidal + Coordinate & $\times$ \\
\hline
\end{tabular}
\label{tab:existing_encodings_comparison}
\end{table}

\begin{table}[t!]
\centering
\caption{Comparison of common geospatial datasets}
\begin{tabular}{l|c|c|c}
\hline
Dataset & Multimodal & Temporal Resolution & Metadata \\
\hline
\textsc{FAIR-Earth} & \checkmark & \checkmark & \checkmark  \\
NaturalEarth & \checkmark & $\times$ & $\times$ \\
\textsc{ERA5} \citep{hersbach1999era5} & $\times$ & \checkmark & \checkmark \\
OpenStreetMap \citep{OpenStreetMap} & $\times$ & \checkmark & \checkmark \\
\hline
\end{tabular}
\label{tab:dataset_comparison}
\end{table}

\clearpage
\newpage
\subsection{Extended Experimental Results}
\label{sec:Experimental-Results}

\begin{table*}[t!]
\centering
\setlength{\tabcolsep}{4pt}
\small

\begin{subfigure}{}
\centering
\begin{tabular}{l|cccc}
    \toprule
    PE $\downarrow$ Resolution $\rightarrow$ & 5000 & 10000 & 15000 & 20000 \\
    \midrule
    \textsc{Spherical Wavelet} & \underline{0.1419 $\pm$ 0.0009} & \underline{0.0890 $\pm$ 0.0003} & \underline{0.1017 $\pm$ 0.0007} & \underline{0.0762 $\pm$ 0.0005} \\
    \textsc{Spherical Harmonic} & 0.1462 $\pm$ 0.0013 & 0.0933 $\pm$ 0.0010 & 0.1057 $\pm$ 0.0010 & 0.0767 $\pm$ 0.0005 \\
    \textsc{Theory} & 0.1586 $\pm$ 0.0034 & 0.1137 $\pm$ 0.0012 & 0.1324 $\pm$ 0.0061 & 0.1016 $\pm$ 0.0073 \\
    \textsc{SphereM+} & 0.6314 $\pm$ 0.0006 & 0.6307 $\pm$ 0.0003 & 0.6358 $\pm$ 0.0001 & 0.6259 $\pm$ 0.0001 \\
    \textsc{SphereC+} & 0.6307 $\pm$ 0.0017 & 0.6308 $\pm$ 0.0002 & 0.6357 $\pm$ 0.0004 & 0.6259 $\pm$ 0.0002 \\
    \bottomrule
\end{tabular}
\caption{\textsc{FAIR-Earth}: Land-Sea Binary Cross-Entropy Loss}
\vspace{6mm}
\end{subfigure}

\vspace{4mm}

\begin{subfigure}{}
\centering
\begin{tabular}{l|cccc}
    \toprule
    PE $\downarrow$ Resolution $\rightarrow$ & 5000 & 10000 & 15000 & 20000 \\
    \midrule
    \textsc{Spherical Wavelet} & 0.0177 $\pm$ 0.0005 & 0.0108 $\pm$ 0.0002 & 0.0095 $\pm$ 0.0001 & 0.0084 $\pm$ 0.0001 \\
    \textsc{Spherical Harmonic} & \underline{0.0158 $\pm$ 0.0005} & \underline{0.0104 $\pm$ 0.0005} & \underline{0.0089 $\pm$ 0.0002} & \underline{0.0078 $\pm$ 0.0003} \\
    \textsc{Theory} & 0.0254 $\pm$ 0.0028 & 0.0144 $\pm$ 0.0016 & 0.0133 $\pm$ 0.0008 & 0.0114 $\pm$ 0.0010 \\
    \textsc{SphereM+} & 2.1551 $\pm$ 0.0043 & 2.2086 $\pm$ 0.0012 & 2.1889 $\pm$ 0.0019 & 2.2089 $\pm$ 0.0015 \\
    \textsc{SphereC+} & 2.1599 $\pm$ 0.0032 & 2.2055 $\pm$ 0.0010 & 2.1856 $\pm$ 0.0018 & 2.2074 $\pm$ 0.0013 \\
    \bottomrule
\end{tabular}
\caption{\textsc{FAIR-Earth}: Carbon Emission Mean-Squared Error}
\vspace{6mm}
\end{subfigure}

\vspace{4mm}

\begin{subfigure}{}
\centering
\begin{tabular}{l|cccc}
    \toprule
    PE $\downarrow$ Resolution $\rightarrow$ & 5000 & 10000 & 15000 & 20000 \\
    \midrule
    \textsc{Spherical Wavelet} & 0.0306 $\pm$ 0.0008 & 0.0237 $\pm$ 0.0001 & 0.0169 $\pm$ 0.0001 & 0.0175 $\pm$ 0.0002 \\
    \textsc{Spherical Harmonic} & \underline{0.0269 $\pm$ 0.0004} & \underline{0.0203 $\pm$ 0.0004} & \underline{0.0158 $\pm$ 0.0003} & \underline{0.0161 $\pm$ 0.0003} \\
    \textsc{Theory} & 0.0535 $\pm$ 0.0060 & 0.0270 $\pm$ 0.0010 & 0.0190 $\pm$ 0.0006 & 0.0194 $\pm$ 0.0023 \\
    \textsc{SphereM+} & 1.9559 $\pm$ 0.0141 & 1.8413 $\pm$ 0.0019 & 1.9528 $\pm$ 0.0026 & 1.8573 $\pm$ 0.0007 \\
    \textsc{SphereC+} & 1.9621 $\pm$ 0.0086 & 1.8433 $\pm$ 0.0025 & 1.9534 $\pm$ 0.0024 & 1.8578 $\pm$ 0.0007 \\
    \bottomrule
\end{tabular}
\caption{\textsc{FAIR-Earth}: Surface Temperature Mean-Squared Error}
\vspace{6mm}
\end{subfigure}

\vspace{4mm}

\begin{subfigure}{}
\centering
\begin{tabular}{l|cccc}
    \toprule
    PE $\downarrow$ Resolution $\rightarrow$ & 5000 & 10000 & 15000 & 20000 \\
    \midrule
    \textsc{Spherical Wavelet} & 0.0993 $\pm$ 0.0004 & \underline{0.0980 $\pm$ 0.0007} & 0.0986 $\pm$ 0.0002 & \underline{0.0976 $\pm$ 0.0006} \\
    \textsc{Spherical Harmonic} & \underline{0.0984 $\pm$ 0.0004} & 0.0984 $\pm$ 0.0006 & \underline{0.0982 $\pm$ 0.0004} & 0.0984 $\pm$ 0.0005 \\
    \textsc{Theory} & 0.1150 $\pm$ 0.0067 & 0.1162 $\pm$ 0.0022 & 0.1195 $\pm$ 0.0014 & 0.1201 $\pm$ 0.0054 \\
    \textsc{SphereM+} & 0.5981 $\pm$ 0.0002 & 0.5979 $\pm$ 0.0001 & 0.5981 $\pm$ 0.0003 & 0.5982 $\pm$ 0.0001 \\
    \textsc{SphereC+} & 0.5981 $\pm$ 0.0001 & 0.5986 $\pm$ 0.0002 & 0.5980 $\pm$ 0.0002 & 0.5983 $\pm$ 0.0002 \\
    \bottomrule
\end{tabular}
\caption{Land-Sea (\cite{rußwurm2024geographiclocationencodingspherical}) Cross-Entropy Loss}
\vspace{6mm}
\end{subfigure}

\vspace{4mm}

\begin{subfigure}{}
\centering
\begin{tabular}{l|cccc}
    \toprule
    PE $\downarrow$ Resolution $\rightarrow$ & 5000 & 10000 & 15000 & 20000 \\
    \midrule
    \textsc{Spherical Wavelet} & \underline{0.1622} $\pm$ 0.0011 & 0.1116 $\pm$ 0.0007 & 0.0848 $\pm$ 0.0010 & 0.0718 $\pm$ 0.0002 \\
    \textsc{Spherical Harmonic} & 0.1631 $\pm$ 0.0011 & \underline{0.1080 $\pm$ 0.0010} & \underline{0.0773 $\pm$ 0.0007} & \underline{0.0652 $\pm$ 0.0011} \\
    \textsc{Theory} & 0.2456 $\pm$ 0.0124 & 0.1796 $\pm$ 0.0068 & 0.1382 $\pm$ 0.0055 & 0.1265 $\pm$ 0.0068 \\
    \textsc{SphereM+} & 1.3853 $\pm$ 0.0003 & 1.3861 $\pm$ 0.0001 & 1.3862 $\pm$ 0.0001 & 1.3862 $\pm$ 0.0000 \\
    \textsc{SphereC+} & 1.3856 $\pm$ 0.0002 & 1.3861 $\pm$ 0.0001 & 1.3863 $\pm$ 0.0000 & 1.3862 $\pm$ 0.0000 \\
    \bottomrule
\end{tabular}
\caption{Checkerboard (\cite{rußwurm2024geographiclocationencodingspherical}) Cross-Entropy Loss}
\vspace{6mm}
\end{subfigure}
\end{table*}

\begin{table}[t!]
    \centering
    \label{tab:encodings}
    \caption{Surface temperature regression subgroup losses for various encodings. Coast (underlined) consistently exhibits greater losses.}
    \begin{tabular}{l|cccc}
        \hline
        Subgroup & \textbf{\textsc{SW}} & \textbf{\textsc{SH}} & \textbf{\textsc{Theory}} & \textbf{\textsc{SphereC+}} \\
        \hline
        Land & 0.087 & 0.076 & 0.217 & 5.867 \\
        Sea & 0.043 & 0.028 & 0.063 & 2.152 \\
        Island & 0.049 & 0.041 & 0.047 & 2.101 \\
        \underline{Coast} & 0.101 & 0.083 & 0.249 & 5.835 \\
        \hline
    \end{tabular}
    \label{tab:subgroup-temperature}
\end{table}

\begin{table*}[t!]
\centering
\setlength{\tabcolsep}{6pt}
\small
\begin{tabular}{l|cc|cc}
    \toprule
    & \multicolumn{2}{c|}{\textbf{Land-Sea Classification}} & \multicolumn{2}{c}{\textbf{Surface Temperature Regression}} \\
    \cmidrule(lr){2-3} \cmidrule(lr){4-5}
    \textbf{Encoding} & \textbf{Best Country} & \textbf{Worst Country} & \textbf{Best Country} & \textbf{Worst Country} \\
    & (Value) & (Value) & (Value) & (Value) \\
    \midrule
    \textsc{Spherical Wavelet} & Finland & Guyana & Denmark & Cambodia \\
    & (0.002) & (0.816) & (0.009) & (0.232) \\
    \midrule
    \textsc{Spherical Harmonic} & Honduras & Spain & Georgia & Panama \\
    & (0.001) & (0.795) & (0.001) & (0.333) \\
    \midrule
    \textsc{Theory} & Kyrgyzstan & Spain & Sierra Leone & Vietnam \\
    & (0.001) & (1.114) & (0.002) & (0.561) \\
    \midrule
    \textsc{SphereC+} & Austria & Chile & Romania & Greenland \\
    & (0.418) & (1.257) & (0.143) & (10.425) \\
    \bottomrule
\end{tabular}
\caption{Country discrepancies via \textsc{FAIR-Earth}. Losses are respective to each dataset, and only countries with over 100 sampled points are included to mitigate noise. All encoding-dataset combinations exhibit a wide disparity in country-level performance.}
\label{tab:encoding-comparison}
\end{table*}

\begin{figure}[t!]
   \centering
   \includegraphics[width=1.0\textwidth]{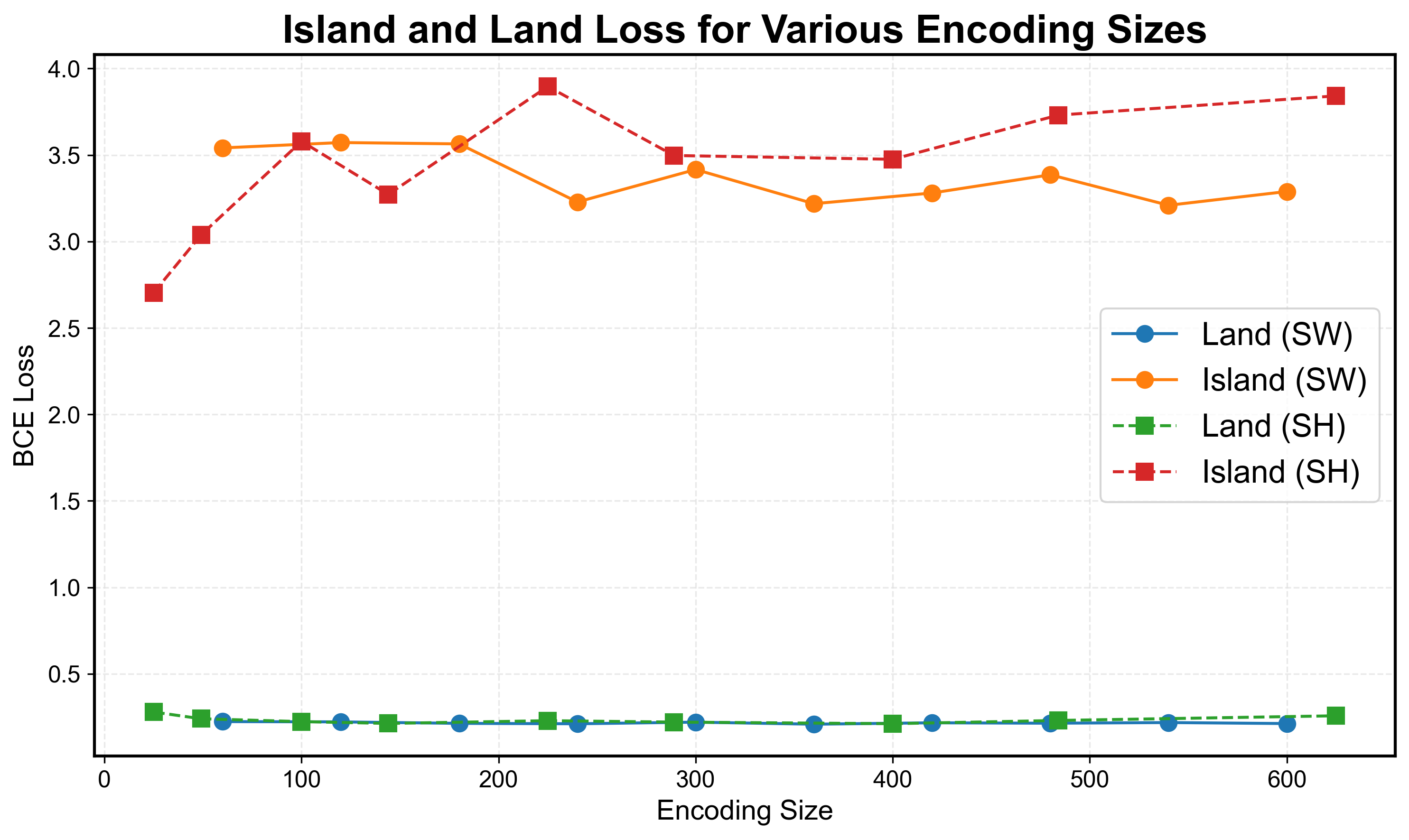}
   \caption{Land and island loss for \textsc{SH}, \textsc{SW} representations, trained on the same resolution with varying encoding sizes. The improved performance of \textsc{Spherical Wavelet} may be attributed to better representation of these fine-scale features. Interestingly, \textsc{Spherical Harmonic}'s biases against islands \textit{worsen} with increased encoding size.}
   \label{fig:sizeAblation}
\end{figure}

\begin{figure}[t!]
   \centering
   \includegraphics[width=0.8\textwidth]{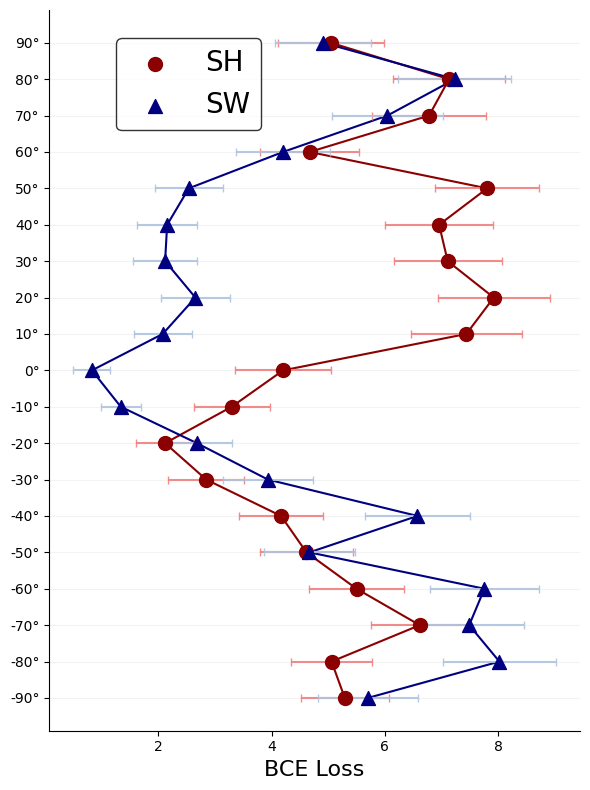}
   \caption{Differences in latitudinal loss on land-sea (\cite{rußwurm2024geographiclocationencodingspherical}) classification, where means and error bars are based off of 1000 uniform samples at each latitude. \textsc{Spherical Wavelet} performance clearly deteriorates as we move from the equator to the poles.}
   \label{fig:latitudeAblation}
\end{figure}

\clearpage
\newpage
\subsection{Wavelet Notes}
\label{sec:wavelet-notes}
\paragraph{Computational Analysis} 
A comparison of the computational requirements for \textsc{Spherical Wavelet} and \textsc{Spherical Harmonic} encodings reveals a discrepancy in computational cost. Starting from our previous formulation, we define a family of valid wavelets as:
$$\{\psi_{a,\rho} \equiv \mathcal{R}(\rho)\mathcal{D}(a)\psi_M\}$$
where the dilation operator $\mathcal{D}(a)$ is defined as:
$[\mathcal{D}(a)s](\omega) = [\lambda(a,\omega)]^{1/2} s(\omega_{1/a})$, with $\lambda$ being a scalar-valued operator.

The rotation operator $\mathcal{R}$ requires only a transformation into Euler space and three $3 \times 3$ matrix multiplications. Since the generation of rotation points is a one-time computation, this yields a computational complexity that scales linearly with encoding size.

In contrast, the closed-form \textsc{Spherical Harmonic} encoding with $L$ Legendre polynomials requires calculation of:
$$\sqrt{\frac{2l+1}{4\pi} \frac{(l-|m|)!}{(l+|m|)!}} P_l^m(\cos \lambda)$$
for all $m \leq l \leq L$. This higher computational burden makes \textsc{Spherical Wavelet} comparatively efficient for larger encoding sizes. Additionally, computational and representational challenges of the factorials lead to instability at higher $L$ values (\ref{fig:legendreNans}), thus we restrict our analyses to $L=30$.

\begin{figure}[t!]
   \centering
   \includegraphics[width=0.8\textwidth]{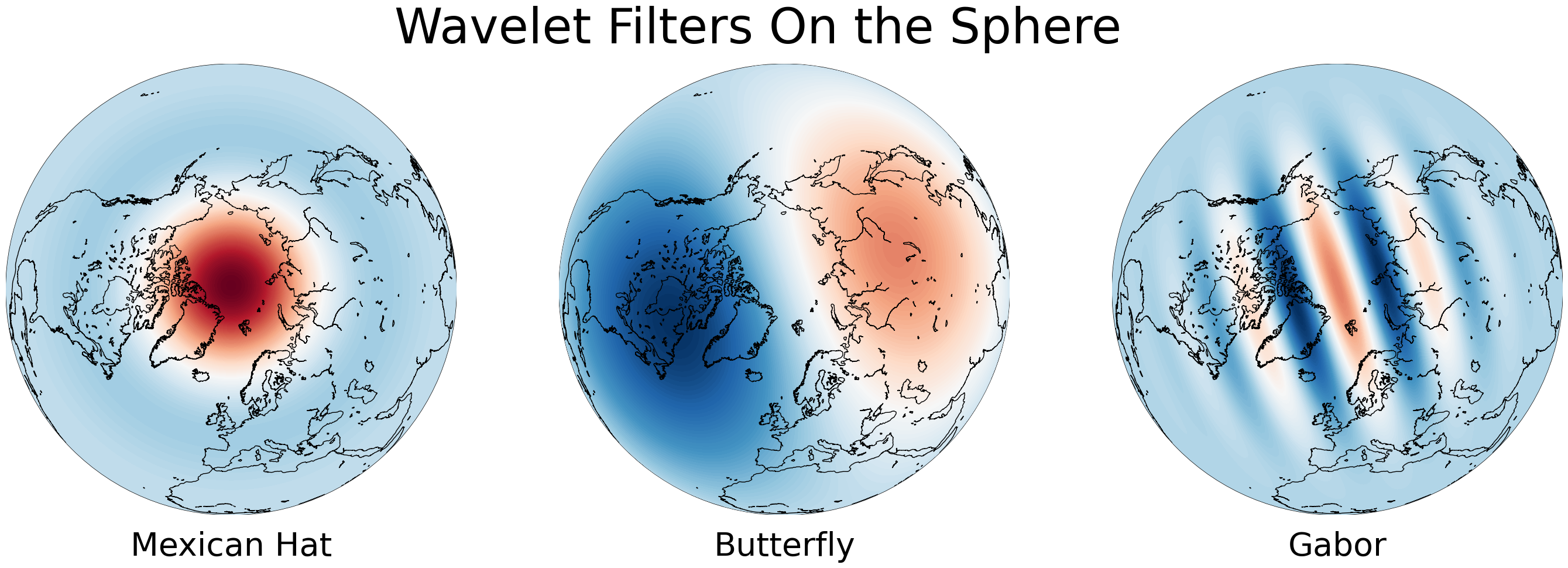}
   \caption{Visualizations of various wavelet filters on the sphere. Of the three, only Gabor consistently reaches state-of-the-art performance.}
   \label{fig:waveletFilters}
\end{figure}

\paragraph{Non-Gabor Filters} We extended our investigation beyond Gabor wavelets to include preliminary experiments with spherical Butterfly and Mexican Hat wavelets, applying the same procedure with different mother wavelets. However, visualizing these wavelets' projections on the sphere suggests they are poorly suited for general Earth signals (\cref{fig:waveletFilters}). In our experiments on the \textsc{FAIR-Earth} land-sea classification task, these alternatives performed worse than Gabor wavelets and even certain baseline encodings when controlling for all other parameters (\cref{fig:wavelet-filter-perf}). Future research directions could explore which signals these filters might be better suited for, and whether wavelet-related adjustments to INRs (\cite{saragadam2023wirewaveletimplicitneural}) could address their current limitations.

\begin{figure}[t!]
\centering
\begin{tabular}{l|cc|cc}
  \toprule
  & \multicolumn{2}{c|}{Land-Sea} & \multicolumn{2}{c}{Surface Temperature} \\
  \cmidrule(lr){2-3} \cmidrule(lr){4-5}
  Filter & 5000 & 10000 & 5000 & 10000 \\
  \midrule
  \underline{Gabor (Ours)} & 0.1419 $\pm$ 0.0009 & 0.0890 $\pm$ 0.0003 & 0.0306 $\pm$ 0.0008 & 0.0237 $\pm$ 0.0001 \\
  Butterfly & 0.1943 $\pm$ 0.0005 & 0.1223 $\pm$ 0.0003 & 0.0523 $\pm$ 0.0005 & 0.0492 $\pm$ 0.0010 \\
  Mexican Hat & 0.4001 $\pm$ 0.0010 & 0.3624 $\pm$ 0.0021 & 0.1550 $\pm$ 0.0018 & 0.1374 $\pm$ 0.0009 \\
  \bottomrule
\end{tabular}
\caption{A comparison of different wavelet filter performance on the \textsc{FAIR-Earth} land-sea classification task. Loss metric is respective to each dataset, and all other parameters are held constant.}
\label{fig:wavelet-filter-perf}
\vspace{6mm}
\end{figure}


\end{document}